\documentclass[10pt,twocolumn,letterpaper]{article}
\usepackage{style/iccv}
\usepackage{times}
\usepackage{epsfig}
\usepackage{rotating, graphicx}
\usepackage{amsmath}
\usepackage{amssymb}
\usepackage{booktabs}
\usepackage{comment}
\usepackage{lipsum}
\usepackage{multirow}
\usepackage{pifont}
\usepackage{subcaption}
\usepackage{scalerel}
\usepackage{enumitem}
\usepackage{soul}
\usepackage{float}
\usepackage[page]{appendix}
\usepackage{multirow, makecell}

\usepackage[skip=5pt,font=footnotesize]{caption}

\newcommand{\tablefontsize}{\small}
\newcommand{\denselist}{\itemsep -2pt\topsep-8pt\partopsep-8pt}

\newcommand{\custompar}[1]{\vspace{-3mm}\paragraph{#1}}

\usepackage[pagebackref=true,breaklinks=true,letterpaper=true,colorlinks,bookmarks=false]{hyperref}
\iccvfinalcopy %

\def\iccvPaperID{9454} %

\ificcvfinal\fi

\definecolor{bluelink}{RGB}{66, 133, 244}

\newcommand{\modelname}{{Deep-MAC }}
\newcommand{\modelnamenospace}{{Deep-MAC}}
\newcommand{\modelnamemaskrcnn}{{Deep-MARC }}
\newcommand{\modelnamemaskrcnnnospace}{Deep-MARC}
\newcommand{\voctrain}{{\tt VOC-Masks-Only}}

\begin{document}

\title{The surprising impact of mask-head architecture on novel class segmentation}

\author{Vighnesh Birodkar,\
 
Zhichao Lu,\ 
Siyang Li,\ 
Vivek Rathod,\ 
Jonathan Huang\\
Google\\
{\tt\small \{vighneshb,lzc,siyang,rathodv,jonathanhuang\}@google.com}
}

\maketitle
\ificcvfinal\thispagestyle{empty}\fi

\begin{abstract}
\vspace{-2mm}
Instance segmentation models today are very accurate when trained on large annotated datasets, but collecting mask annotations at scale is prohibitively expensive. We address the partially supervised instance segmentation problem in which one can train on (significantly cheaper) bounding boxes for all categories but use masks only for a subset of categories. 
In this work, we focus on a popular family of models which apply differentiable cropping to a feature map
and predict a mask based on the resulting crop. %
Under this family, we study Mask R-CNN and discover
that instead of its default strategy of training 
the mask-head with a combination of proposals and groundtruth
boxes, training the mask-head with only groundtruth
boxes dramatically improves its performance
on novel classes.
This training strategy also allows us to take
advantage of alternative mask-head architectures,
which we exploit by replacing
the typical mask-head of 2-4 layers
with significantly deeper off-the-shelf architectures 
(e.g. ResNet, Hourglass models).
While many of these architectures perform
similarly when trained in fully supervised mode, our main finding is
that they can generalize to novel classes in dramatically
different ways.
We call this ability of mask-heads to generalize
to unseen classes the 
\emph{strong mask generalization} effect
and show that without any specialty
modules or losses, we can achieve
state-of-the-art results in the partially
supervised COCO instance segmentation benchmark.
Finally, we demonstrate that our effect is general, holding across
underlying detection methodologies (including anchor-based, anchor-free or no detector at all) and across different backbone networks.
Code and pre-trained models are available at \url{ https://git.io/deepmac}.

\end{abstract}

\vspace{-3mm}
\section{Introduction}\vspace{-2mm}

\begin{figure}
\centering

\setlength{\fboxsep}{0pt}%
\setlength{\fboxrule}{0.5pt}
\begin{subfigure}[t]{0.24\linewidth}
\centering
\fbox{\includegraphics[width=0.9\linewidth]{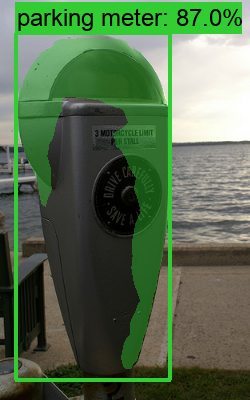}}

\vspace{5pt}

\fbox{\includegraphics[trim=90 0 60 0,clip,width=0.9\linewidth]{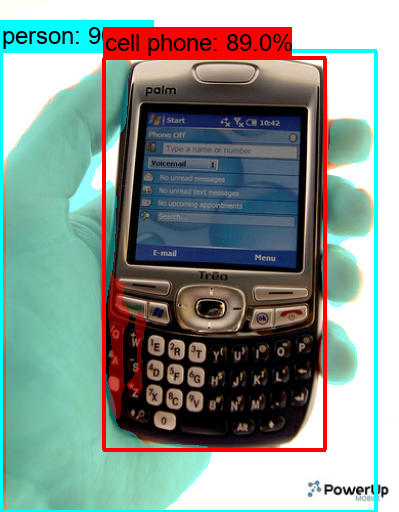}}

\vspace{5pt}

\fbox{\includegraphics[trim=150 0 20 0,clip, width=0.9\linewidth]{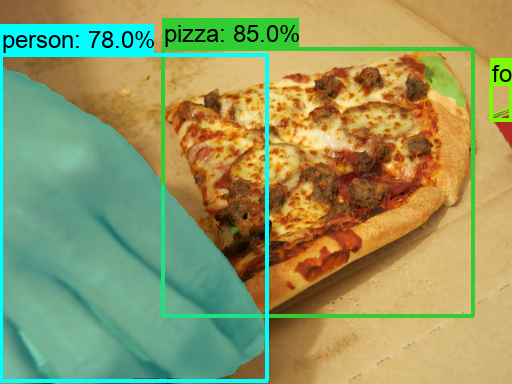}}

\caption{ResNet-4}
\end{subfigure}
\begin{subfigure}[t]{0.24\linewidth}
\centering
\fbox{\includegraphics[width=0.9\linewidth]{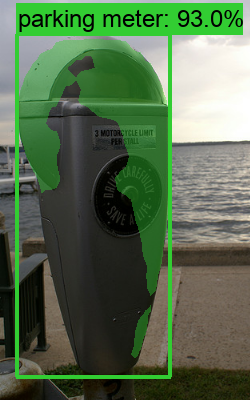}}

\vspace{5pt}

\fbox{\includegraphics[trim=90 0 60 0,clip,width=0.9\linewidth]{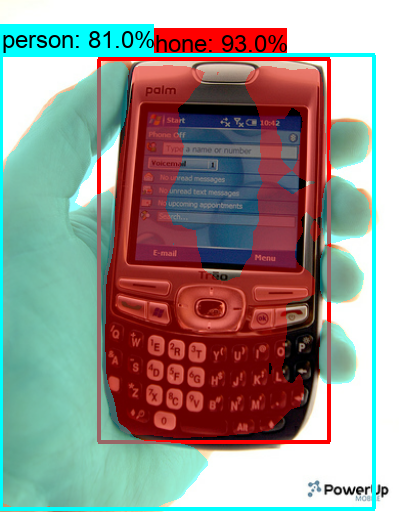}}

\vspace{5pt}

\fbox{\includegraphics[trim=150 0 20 0,clip, width=0.9\linewidth]{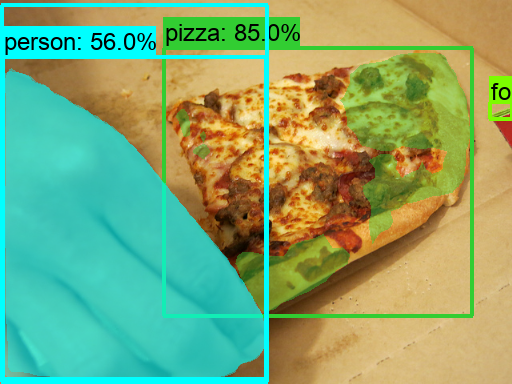}}
\caption{ResNet-12}
\end{subfigure}
\begin{subfigure}[t]{0.24\linewidth}
\centering
\fbox{\includegraphics[width=0.9\linewidth]{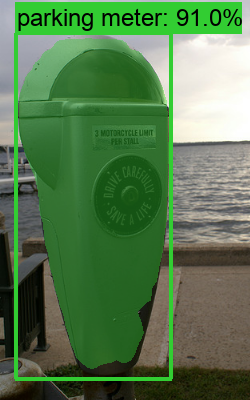}}

\vspace{5pt}

\fbox{\includegraphics[trim=90 0 60 0,clip,width=0.9\linewidth]{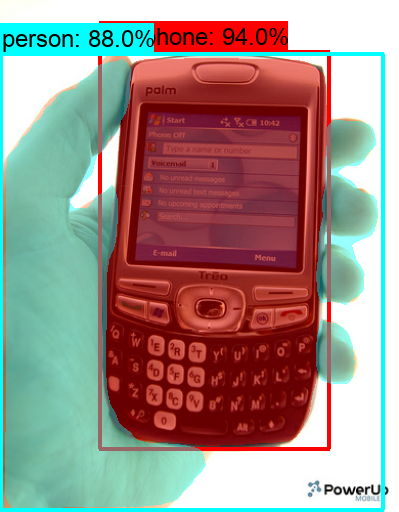}}

\vspace{5pt}

\fbox{\includegraphics[trim=150 0 20 0,clip, width=0.9\linewidth]{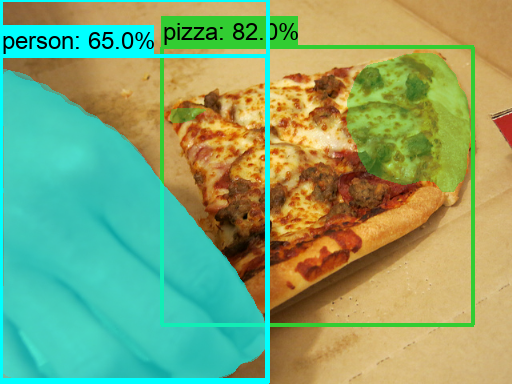}}
\caption{ResNet-20}
\end{subfigure}
\begin{subfigure}[t]{0.24\linewidth}
\centering
\fbox{\includegraphics[width=0.9\linewidth]{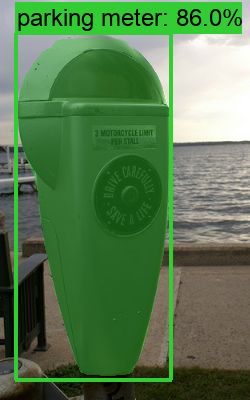}}

\vspace{5pt}

\fbox{\includegraphics[trim=90 0 60 0,clip,width=0.9\linewidth]{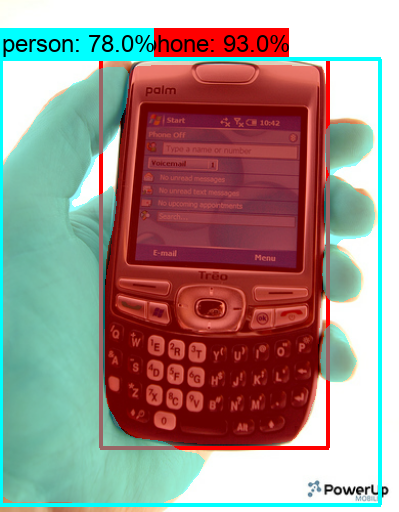}}

\vspace{5pt}

\fbox{\includegraphics[trim=150 0 20 0,clip, width=0.9\linewidth]{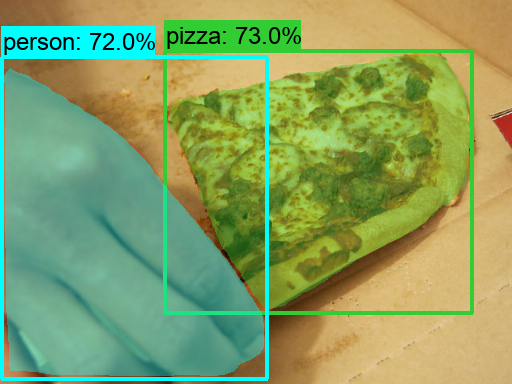}}
\caption{Hourglass-20}
\end{subfigure}
\caption{
The effect of mask-head architecture on mask predictions for unseen
classes.
\emph{Despite never having seen masks from the `parking meter', `pizza' or
`mobile phone' class}, the rightmost mask-head architecture can segment
these classes correctly. 
From left to right, we show better mask-head architectures predicting
better masks.  Moreover, this difference is \emph{only} apparent when evaluating on 
unseen classes --- if we evaluate on seen classes, all four architectures 
exhibit similar performance.}\vspace{-3mm}
\label{fig:prettypicture}
\end{figure}

Large labeled datasets like COCO \cite{coco} %
are crucial for deep neural network based
instance segmentation methods \cite{maskrcnn,cascade,detectors}.
However, collecting groundtruth masks can take $>10\times$ more time
than bounding box annotations.
In COCO \cite{coco}, mask annotations required $\approx 80$ seconds
on average %
whereas methods such as Extreme Clicking \cite{xclick} yield bounding boxes
in 7 seconds.

Given that boxes are much cheaper to annotate than masks,
we address the ``partially supervised'' instance segmentation problem \cite{segment_everything},
where all classes have bounding box annotations but only a subset of
classes have  mask annotations.  We will refer to classes with mask annotations as ``seen'' categories
and classes without as ``unseen''.
Doing well on this task requires the model to generalize in a strong sense,
producing correct masks on unseen classes.

We consider a general family of \emph{crop-then-segment} instance segmentation models where one extracts a feature map over an image, then given a tight bounding box around an instance, performs a differentiable crop
(e.g. ROIAlign~\cite{maskrcnn}). The cropped feature map is then fed to a mask-head subnetwork 
to yield a final mask prediction.  This mask prediction is performed in a class-agnostic manner
so that a model trained from a subset of classes can be applied unchanged to novel classes.

One ``na\"{i}ve'' baseline in this family is to adapt Mask R-CNN~\cite{maskrcnn} to produce class-agnostic masks.
But this approach is known to perform abysmally on unseen classes
(e.g. on the standard partially supervised COCO benchmark, it achieves
$<20\%$ mask mAP on unseen classes vs $>40\%$ on seen, ~\cite{segment_everything}).
Thus previous approaches have used, e.g., offline-trained shape priors~\cite{shapemask} or specialty losses~\cite{cpnet} yielding significantly improved results.

As a starting point, we revisit ``na\"{i}ve'' Mask R-CNN to better understand the reasons for its poor performance.  Our first finding is that the typical strategy of training the Mask R-CNN mask-head with a combination
of groundtruth and proposed (typically noisy) boxes is a major culprit 
that inhibits its performance on novel classes.
While training with noisy proposals gives slightly better results when
fully supervised,
we show that simply training the mask-head with \emph{only}
groundtruth boxes has a surprising impact on its performance
on unseen classes (+9 mAP) (note that we follow the usual procedure of using predicted boxes at test time).

We next zoom out beyond Mask R-CNN to the more general family of crop-then-segment models. Our second major finding is that in the context of using the above slightly modified training regime, the architecture of the mask-head takes on a disproportionately impactful role in generalization to unseen classes.  
More specifically, we find that 
mask-heads that might perform similarly under full supervision can behave
differently under partial supervision, generalizing to unseen classes in strikingly different ways.

While it is natural to experiment with different
mask-head architectures, we note that their role in generalization
has not been studied extensively in prior literature likely for the following reasons: (1) the choice of mask-head architecture has limited impact in the fully supervised
setting, (2) heavier mask-heads adversely impact running time.
and (3) as noted above, in  architectures like Mask R-CNN, the benefits
of using better mask-heads are not necessarily realized 
 in the default training regimen.
 Thus most prior works in instance segmentation
have settled on using shallow (2-4 layer) fully connected or
convolution based mask-heads.

In our COCO experiments, we find that the difference between worst and best architectures is only ~1\% (absolute mAP) on seen classes but can be ~7\% on unseen classes (examples in Figure~\ref{fig:prettypicture}).
This difference is visually palpable and subsequently changes the calculus for deciding whether it's worth using a heavier mask-head. 

We refer to this effect of certain mask-head architectures on unseen classes as the \emph{``strong mask generalization effect''} and illustrate it with 3 representative model classes: an anchor-free and anchor-based model, and one that discards detection altogether.  
We show that our effect is general, holding across underlying detection methodologies (or no detector at all)
and across different backbone networks.
We also identify architectural characteristics (such as depth and encoder-decoder arrangements)
that empirically yield strong mask generalization properties. 

One main finding is that deeper mask-heads generalize better 
despite being counter-intuitively more over-parameterized than shallower ones.
Our anchor-based model, based on Mask R-CNN~\cite{maskrcnn},
employs mask-heads that are 20+ layers deep and we thus refer
to this model as \modelnamemaskrcnnnospace~(for Deep Mask-heads Above R-CNN).
Similarly, our anchor-free model, which we use for most ablations,
is based on CenterNet~\cite{centernet} and is called
\modelnamenospace~(for Deep Mask-heads Above CenterNet).
Using out-of-the-box mask-head architectures, we show
that both \modelname and \modelnamemaskrcnn surpass
the state-of-the-art \cite{cpnet} in the COCO partially
supervised instance segmentation setting with 35.5 \% and 38.7 \% mAP respectively.

Due to space limitations, we have relegated a number of 
auxiliary findings to the Appendix.  Among them, we show that:
(1) two-stage training (i.e. self-distillation) helps, allowing us to achieve 40.4\% mask mAP on unseen categories (Section~\ref{sec:distiallation}); (2) our models have reached a likely
saturation point in terms of mask quality on COCO
(Section~\ref{sec:limited_headroom}) --- the implication is that future improvements on this particular benchmark are far more likely to come from detection; and (3) we demonstrate that we can achieve surprisingly strong mask
generalization results with just 1 seen class (depending on the class, Section~\ref{sec:single_class}).

{\setlength{\parindent}{0cm}
We summarize our main contributions as follows:
\vspace{-2mm}
\begin{itemize}[]\denselist
  \item We identify the strong mask generalization effect in
  partially supervised instance segmentation %
  and show that it is general, holding across underlying detectors like 
  Mask
  R-CNN~\cite{maskrcnn} and CenterNet~\cite{centernet} or without a detector, and across different backbones (Section ~\ref{sec:going_deeper_mask_heads}).
  \item In order to unlock strong mask generalization, we show that it is
  necessary to train using tight groundtruth boxes instead of a combination
  of groundtruth and noisy proposals.  We revisit vanilla Mask R-CNN
  with this insight and show that this change alone 
  dramatically improves the performance 
  on unseen classes %
  (Section~\ref{sec:importance_gt}).
  \item We identify characteristics of mask-head architectures that lead to strong mask generalization (Section~\ref{sec:ablations}). Among other things, we find that Hourglass~\cite{hourglass} architectures offer excellent performance.
  We use these findings to achieve state-of-the-art results on the COCO partially supervised instance segmentation task (Section~\ref{sec:sota}) with our CenterNet and Mask R-CNN
  based models, \modelname and \modelnamemaskrcnn.
\end{itemize}
}

\section{Related work}\label{sec:related}
\vspace{-1mm}
\paragraph{Object detection and instance segmentation.}
There has been a significant progress over the last decade in detection with successful convolutional models like OverFeat~\cite{sermanet2013overfeat}, YOLO~\cite{redmon2016you,redmon2017yolo9000,redmon2018yolov3,bochkovskiy2020yolov4}, Multibox~\cite{szegedy2013deep,erhan2014scalable}, SSD~\cite{liu2016ssd}, RetinaNet~\cite{lin2017focal}, R-CNN and Fast/Faster versions~\cite{girshick2015region,girshick2015fast,ren2015faster,he2016deep}, EfficientDet~\cite{tan2020efficientdet}, etc. 
While many of these works initially focused on box detection, more recently, many benchmarks have focused on the more detailed problem of instance segmentation (COCO~\cite{coco}, OID v5~\cite{kuznetsova2018open,benenson2019large}, LVIS~\cite{gupta2019lvis}) and panoptic segmentation (COCO-Panoptic,~\cite{kirillov2019panoptic}) which are arguably more useful tasks in certain applications.  
A major milestone in this literature was  Mask R-CNN~\cite{maskrcnn} which influenced many SOTA approaches today (e.g.,~\cite{detectors,liu2018path}) and 
by itself continues to serve as a strong baseline.
\custompar{Anchor-free methods.}
State-of-the-art methods today are predominantly built on anchor-based approaches which predict classification/box offsets relative to a collection of fixed boxes arranged in sliding window fashion (called ``anchors'').  While effective, the performance of anchor-based methods often depend on manually-specified design decisions, e.g. anchor layouts and target assignment heuristics, a complex space to navigate for practitioners. 

In recent years, however, this monopoly has been broken with the introduction of competitive ``anchor-free'' approaches~\cite{law2018cornernet,duan2019centernet,tian2019fcos,extremenet,centernet,kong2020foveabox,zhu2019feature,carion2020end}.  These newer anchor-free methods are simpler, more amenable to extension,
offer competitive performance and consequently are beginning to be popular. %
Our anchor-free model (Section~\ref{sec:model_families}, \modelname) in particular builds on the ``CenterNet''  
architecture~\cite{centernet}.

Due to the recency of competitive anchor-free methods 
there are fewer anchor-free instance segmentation approaches in literature. 
\cite{lee2020centermask,xie2020polarmask,ying2019embedmask,cpnet}
all add mask prediction capabilities on top of the (anchor-free) FCOS~\cite{tian2019fcos} framework.
While the primary focus of our work is partial supervision, the fully supervised version of our model adds to this growing body of work, offering strong performance among anchor-free instance segmentation approaches.

\custompar{Box-only supervision for instance segmentation.}
The above  methods rely on  access to massive labeled datasets
which are costly to develop, with mask annotations  especially so compared to box annotations.
Researchers have thus begun to develop methods that are less reliant on mask annotations.  
In one formulation of this problem (which we might call \emph{strictly box-supervised}) we ask to learn an instance segmentation model given only box annotations and no masks~\cite{khoreva2017simple,remez2018learning,hsu2019weakly,kervadec2020bounding,tian2020boxinst}.  
However this is intuitively a  difficult approach and the performance of all of these methods is still a far cry from fully supervised performance of a strong baseline particularly at high IOU thresholds for mAP.
\custompar{Partial supervision for instance segmentation}
Instead of going to the extreme end of discarding all mask annotations, Hu et al.~\cite{segment_everything} introduced the \emph{partial supervision} formulation which allows for mask annotations from a small subset of classes to be used along with all box annotations.  \cite{segment_everything} observed that the ``obvious’’ baseline of using a class-agnostic version of Mask R-CNN yielded poor results and proposed a method (Mask$^X$) for learning to predict mask-head weights given box-head weights hoping that this learned function will generalize to classes whose masks are not observed at training time.

Later papers~\cite{shapemask,cpnet} however revisited the approach of attaching a class-agnostic mask-head on top of a detector, in both cases introducing novel architectures and additional losses to significantly improve generalization to novel classes.  
ShapeMask~\cite{shapemask} builds on RetinaNet, learning a low dimensional shape space from observed masks and uses projections to this space to guide mask estimation; they also introduce a simple method to ``condition’’ features cropped from the backbone on the instance that is being segmented.
CP-Net~\cite{cpnet}, which is the current state of the art on this problem builds on FCOS~\cite{tian2019fcos}, adding boundary prediction  and attention-based aggregation in the mask branch.

We take a similar approach of using a class-agnostic mask-head, but
while the ideas explored in these prior works are clearly beneficial, our objective is to demonstrate 
that mask-head architecture itself plays an underappreciated but significant role in generalization.
Notably, by exploiting out-of-the-box architectures with strong mask generalization properties, we show that 
with only minor tweaks to the training procedure (Sections~\ref{sec:importance_gt},
\ref{sec:going_deeper_mask_heads}) both of our models, \modelname and
 \modelnamemaskrcnn %
have state of the art performance in the partial supervision task.

\section{Crop-then-segment instance segmentation}
\label{sec:model_families}
\begin{figure}[t!]
    \centering
    \includegraphics[width=0.45\textwidth]{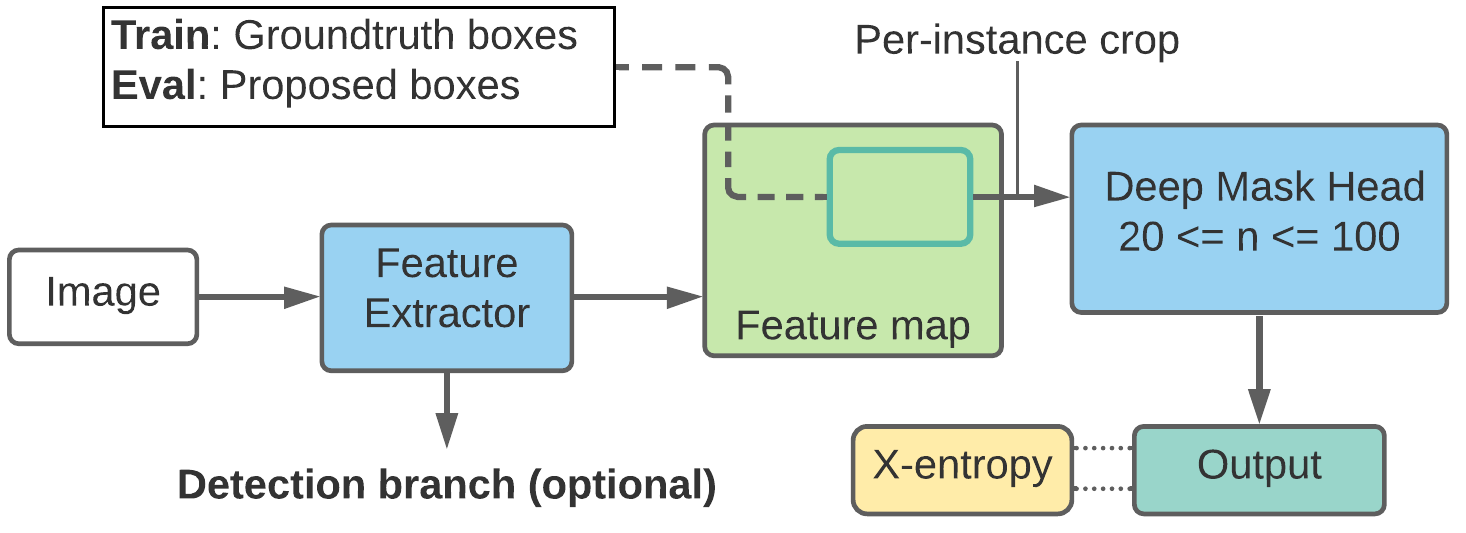}
    \caption{Diagram of the crop-then-segment instance segmentation model family.
    We identify the two crucial features that lead
    to strong mask generalization:
    Using a deeper class-agnostic mask-head
    and training it with \emph{only}
    groundtruth boxes. At test-time, we use predicted boxes. The detection branch is marked
    optional because we show in Section~\ref{sec:detectionfree} that
    it's not required for strong mask generalization.
    }\vspace{-2mm}
    \label{fig:diagram_concept}
\end{figure}

In this paper 
we consider a general family of ``crop-then-segment'' 
models that apply
a per-instance crop (RoIAlign,~\cite{maskrcnn}) operation after a feature extractor and pass
the cropped features to a class-agnostic mask-head. %
For example, in our experiments, we use two detection-based instances of this family building on Mask R-CNN (anchor based) and Centernet (anchor-free), as well as a model that does not perform detection (and is simply provided with bounding boxes as input at test time). A schematic representation of this model family is drawn in Figure \ref{fig:diagram_concept}.

We focus specifically on two choices that one can make for models within the crop-then-segment family: (1) whether to crop to groundtruth boxes or both groundtruth boxes and proposals when training the mask heads (of the detector based models), and (2), which mask-head architecture to use. As we show, in order to achieve strong mask generalization, it is critical to (1) train with only groundtruth and (2) use significantly deeper mask head architectures than what is commonly used. To emphasize these aspects, we refer to our modified detection based models as \emph{\modelnamemaskrcnn (Deep Mask-heads Above R-CNN)} and \emph{\modelname (Deep Mask-heads Above CenterNet)}.  

In both cases we keep the detection part of our models unchanged
from the standard implementation and make
only minimal changes
where required to be compatible with our mask
head architectures. 
Below we discuss our modifications to Mask R-CNN and CenterNet more in detail.

\custompar{\modelnamemaskrcnnnospace: a Mask R-CNN based model.}
\modelnamemaskrcnn is based on a class-agnostic
version of
Mask R-CNN \cite{maskrcnn} %
where we crop to only groundtruth boxes at training time (as mentioned above) and make minor changes to the mask prediction branch of Mask R-CNN, leaving the detection branch untouched.

Mask R-CNN by default crops its feature maps (using RoIAlign) to $14 \times 14$ resolution
and upsamples to $28 \times 28$ before
predicting per-instance masks.
At test-time these are re-aligned with respect to the original box and resized to the resolution of the original image.
When evaluating Mask R-CNN with its default mask-head,
we keep this pathway untouched. Our implementation of Hourglass (HG) networks however requires its
input size to be of the form $2^n$ due to its successive down-sampling and up-sampling layers. 
For our HG-20 mask-head we crop feature maps to $16 \times 16$
and upsample  to $32 \times 32$ before predicting a class-agnostic mask.
For our HG-52 mask-head, the crop and output size is doubled to $32 \times 32$
and $64 \times 64$ respectively. For \modelnamemaskrcnnnospace, we do not use any additional inputs to the mask-head.

\custompar{\modelname: an anchor-free model.}
Our \modelname architecture
builds instance segmentation capabilities
on top of  
CenterNet~\cite{centernet},\footnote{Not to be 
confused for the CenterNet from Duan et al.~\cite{duan2019centernet}.}
a popular anchor-free detection approach, %
which models objects relative to their centers. For predicting
bounding boxes, CenterNet outputs 3 tensors: (1) a class-specific heatmap which
indicates the probability of the center of a bounding box being present at each location,
(2) a class-agnostic 2-channel tensor indicating the height and width of the bounding
box at each center pixel, and
(3) since the output feature map is typically smaller than the image (stride 4 or 8),
CenterNet also predicts an $x$ and $y$ direction offset to recover this discretization
error at each center pixel.

\custompar{Predicting instance masks with CenterNet (\modelnamenospace).}
In parallel to the box-related prediction heads, we add a fourth 
\emph{pixel embedding} branch $P$. %
For each bounding box $b$, we crop a region $P_b$ from $P$ corresponding
to $b$ via ROIAlign~\cite{maskrcnn} %
which results in a $32 \times 32$ tensor. We then  feed each $P_b$ to a mask-head  whose architecture is discussed in Section~\ref{sec:going_deeper_mask_heads}.  
Our final prediction at the end is a class-agnostic $32\times 32$ tensor which 
we pass through a sigmoid to get per-pixel probabilities.  We train this mask-head via a
 per-pixel cross-entropy loss averaged over all pixels and instances.
During post-processing, the predicted mask is re-aligned according to the predicted
box  and resized to the resolution of the image.

In addition to this $32\times 32$ cropped feature map, we add two inputs for improved 
stability of some mask-heads
(but note that our main findings \emph{do not depend} on having these additional inputs;
see Appendix~\ref{sec:emebdding_ablations}): {\bf(1) Instance embedding:} We add an additional
head to the backbone that predicts a per-pixel embedding. For each
bounding box $b$ we extract its embedding from the center pixel. This embedding
is tiled to a size of $32 \times 32$ and concatenated to the pixel embedding crop.
This helps condition the mask-head on a particular instance and disambiguate it from others.
{\bf(2) Coordinate Embedding:} Inspired by CoordConv~\cite{coordconv}, 
    we add a  $32 \times 32 \times 2$ tensor holding normalized
$(x, y)$ coordinates relative to the bounding box $b$.

\section{Experimental Setup}

For all experiments in this paper
we follow the typical partially supervised experimental setup
for the COCO dataset with the 20 Pascal VOC~\cite{pascal} categories having
instance masks at training time 
(as the \emph{seen} categories) and the remaining 60 non-VOC classes not
having instance masks at training time (as the \emph{unseen} categories).
In this case, we mainly care about performance 
on the 60 unseen (Non-VOC)
categories since it is more challenging than
the opposite variant and use this setting by default.
We denote this training setting as 
\texttt{VOC-Masks-Only}. 
The only exception is Table \ref{tab:highres}
where we evaluate both variants to compare with
other methods. All evaluations
are performed on the \texttt{coco-val2017} set.

We train all mask heads with sigmoid cross entropy, and 
to handle the partially annotated training data, mask loss for each instance is only considered if a groundtruth mask
is available.
Below, we discuss  experimental details specific to \modelnamemaskrcnn and \modelnamenospace.
For reference, the fully supervised performance on COCO for \modelname is 39.4 mAP and for \modelnamemaskrcnn is 42.8 mAP
on \texttt{coco-testdev2017} (see Appendix for details, Table \ref{tab:full_super}),
which is competitive with ShapeMask~\cite{shapemask} and CPMask~\cite{cpnet}.

\custompar{\modelnamemaskrcnnnospace.}
We train \modelnamemaskrcnn with ResNet~\cite{resnet} backbones at $1024\times1024$ resolution 
with the $3\times$
schedule from Detectron2~\cite{detectron2}. When using the SpineNet~\cite{spinenet}
backbone we train at $1280 \times 1280$ resolution and use
``Protocol C''~\cite{spinenet}.
The ResNet backbones are initialized from
an ImageNet checkpoint whereas the SpineNet models
are trained from scratch. All our models use synchronized 
batch-normalization~\cite{ioffe2015batch,he2019rethinking}.
\modelnamemaskrcnn is implemented in the TF Vision
API~\cite{tfvision}. We only alter the implementation
of Mask R-CNN to support training with groundtruth boxes.
All other detection and optimization hyperparameters are
kept unchanged from their defaults.

\custompar{\modelnamenospace.}
We use a pixel embedding layer with 16 channels and an instance embedding layer with 32 channels.
For all of our models, we use a mask loss weight of $5.0$
and train with synchronized batch-normalization.
We use the Hourglass-104~\cite{hourglass}
backbone for experiments, unless noted otherwise.
Our best models which beat state-of-the-art
(Section \ref{sec:sota}) are trained at $1024 \times 1024$ resolution, with weights
initialized from a COCO detection checkpoint.
All other models are trained at $512 \times 512$ and 
initialized from an ExtremeNet \cite{extremenet} checkpoint inline with the
original implementation of CenterNet~\cite{centernet}.
For our best results we use CenterNet with the Hourglass~\cite{hourglass}
backbone.
\modelname is built on top of the open-source CenterNet implementation
in the TF Object Detection API~\cite{odapi}.
All other detection and optimization hyperparameters are
kept unchanged from their defaults.

\section{Cropping to \emph{only} groundtruth boxes}
\label{sec:importance_gt}
\begin{table}[t!]
\tablefontsize
\centering
\begin{tabular}{llrrr}
\toprule
 {\bf M.H. Train} & {\bf Resnet} & \multicolumn{3}{c}{\bf Mask mAP} \\
\cmidrule{3-5} 
&& {\bf Overall} & \ \ {\bf VOC} & {\bf Non-VOC} \\
\midrule
Prop.+GT & 50 & 23.5 & 39.5 & 18.2\\
GT-Only & 50 & 29.4 & 39.7 & 25.9\\
Prop.+GT & 101 & 24.9 & 40.9 & 19.6\\
GT-Only & 101 & 32.2 & 41.1 &  29.3\\
\bottomrule
\end{tabular}
\caption{
Impact of Mask R-CNN mask-head training (M.H. Train) strategies
on generalization to unseen classes with
Resnet-50-FPN and Resnet-101-FPN backbones. 
All results are reported with the
\voctrain~ setting.
There is a dramatic improvement
in the performance on unseen classes (Non-VOC)
when we train the mask-head with only groundtruth boxes.
When evaluating, we use predicted boxes. 
}\vspace{-1mm}
\label{tab:importance_gt}
\end{table}

Mask R-CNN is typically trained by performing ROIAlign on a combination of groundtruth boxes and proposals — this is a natural approach as it allows for the training distribution to be statistically more similar to the test time distribution and can even be thought of as a form of data augmentation. And for full supervision setups indeed it is slightly better to train with both groundtruth boxes and proposals (e.g. Table~\ref{tab:maskrcnn_supervised}, Appendix). Our first surprising finding is that this situation is dramatically reversed on unseen classes in partially supervised setups, where we find that it is \emph{far} better to train with only groundtruth boxes.

This effect is illustrated in Table~\ref{tab:importance_gt}, where we see that ``naive’’ Mask R-CNN (\textbf{Prop+GT} rows) achieves extremely low mAP on unseen classes (relative, e.g, to seen classes), which is consistent with previous literature \cite{segment_everything}. Training with groundtruth only (\textbf{GT-only} rows) on the other hand, dramatically improves performance for the Non-VOC (unseen) classes for  which  we  do not  provide  masks  at  training  time  (+7.7mAP and + 9.7 mAP). Note  that even with \textbf{GT-only} training, evaluation is always done with proposed boxes,  as with  all  other  methods  we  compare  with. 

Thus for the remainder of the paper we train only with groundtruth boxes unless otherwise specified.  Why would it help so much to train with groundtruth boxes only?  Our hypothesis (in Section~\ref{sec:mask_head_arch_ablations}) hinges on the finding of the next Section where we see that when training with groundtruth boxes only, mask-head architectures take on a new and  significant role in generalization.

\section{Going deeper with mask heads}
\label{sec:going_deeper_mask_heads}

In this section we pull our second lever by varying mask-head architectures. 
Our main finding is that \emph{mask-heads affect generalization  on unseen classes to
a surprising extent}.
In our experiments, we set our mask-heads to be
Hourglass~\cite{hourglass} (HG) and
Resnets~\cite{resnet} (basic and bottleneck variants), with varying depth.
We also use ResNet bottleneck [1/4$^{\mathrm{th}}$], a variant of the 
ResNet (bottleneck) mask-head with $4\times$ fewer channels.
We set the number of channels in the first layer of all mask-heads to
64, increasing this gradually in deeper layers
(see Appendix, Section~\ref{sec:mask_head_details}).
We also set the number of convolutions of each dimensionality %
roughly similar between mask-heads of similar depth.

\custompar{\modelnamemaskrcnnnospace.}
To begin, let’s continue with our Mask R-CNN based models (henceforth, \modelnamemaskrcnnnospace).  In Table~\ref{tab:deepmarc_mask_heads} we train ResNet-101-FPN based \modelnamemaskrcnn models comparing the default mask-head (comprising 4 convolution layers) 
against the above out-of-the-box architectures (ResNet-4, HG-20, HG-52)
and report mask mAP on seen and unseen classes.  
We first observe that when training with groundtruth boxes,
the mAP on seen classes depends a little bit on the specific mask-head
architecture, but the difference between worst and best case is relatively small ($40.3 \rightarrow 41.9$).
However, for the same settings, the mAP on unseen classes varies much more significantly ($27.4 \rightarrow 34.4$).
This  indicates  that  mask-head  architectures play a critical role in generalization to unseen classes, and not just by virtue of fitting the training classes better.
In fact, using an HG-52 mask-head without proposals is enough for \modelnamemaskrcnn to surpass the previous SOTA~\cite{cpnet}.

\begin{table}
\tablefontsize
\centering
\begin{tabular}{lrrcrr}
\toprule
{\bf Mask-Head} & \multicolumn{2}{c}{\bf VOC mAP} & &\multicolumn{2}{c}{\bf Non-VOC mAP}\\
\cmidrule{2-3} \cmidrule{5-6}
& {\bf Prop. + GT.} & {\bf GT. } && {\bf Prop. + GT.} & {\bf GT.} \\
\midrule
Default & 40.9 & 41.1 && 19.6 & 29.3\\
ResNet-4 & 39.2 & 40.3& & 21.0 & 27.4\\
HG-20 & 41.6 & 41.4& & 20.6 & 33.8\\
HG-52& 42.0 & 41.9& & 20.6 & {\bf 34.4}\\
\bottomrule
\end{tabular}
\caption{
Performance of \modelnamemaskrcnn  with different mask-heads under
the \voctrain~setting with a ResNet-101-FPN backbone, comparing
the performance when training the proposed boxes and
groundtruth boxes (\textbf{Prop.+GT.}) and only groundtruth boxes (\textbf{GT.}).  
We see that performance on unseen classes depends significantly on the mask-head, 
but the benefit of better mask-heads is only
apparent %
when training with groundtruth boxes.
With the Hourglass (HG-52) mask-head and no other bells or whistles,
\modelnamemaskrcnn surpasses the previous state-of-the-art~\cite{cpnet}.
}\vspace{-2mm}
\label{tab:deepmarc_mask_heads}
\end{table}

To circle back to the previous section, we  also
see that this effect is tied to our choice to train with only groundtruth boxes --- 
if we include proposals at training time, our models fare significantly worse on unseen classes 
and there is no clear signal on what mask-head architecture is best.  

\begin{figure}
\begin{center}
\includegraphics[width=1\linewidth]{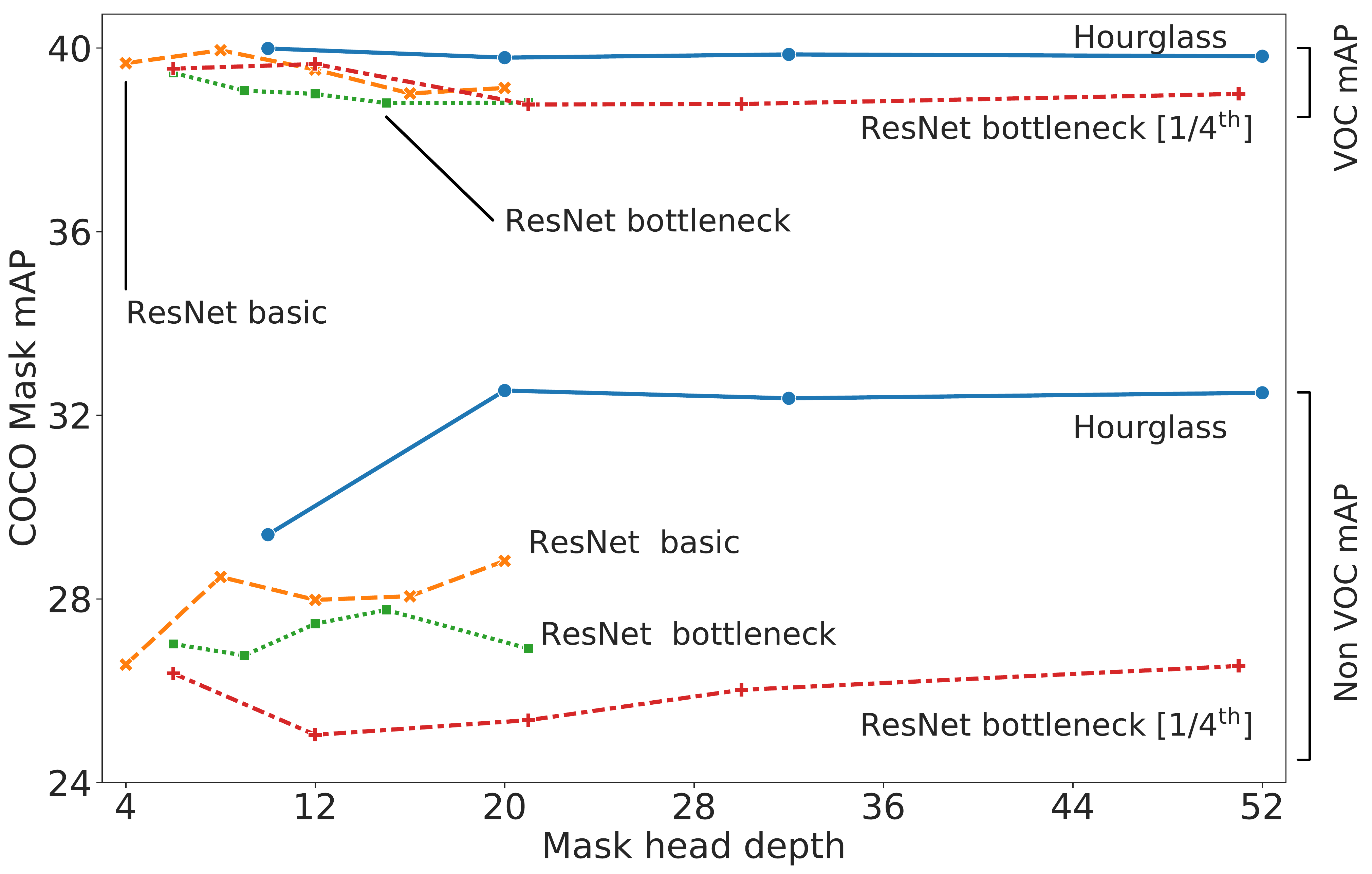}\vspace{-3mm}
\end{center}
\caption{ Effect of mask-head architecture and depth on instance segmentation performance
over seen (VOC) and unseen (Non-VOC) classes. Although the performance on seen classes
does not vary much across different architectures, 
there is significant variation in the performance on unseen classes. We report results with the 
\voctrain~ setting.
}\vspace{-1mm}
\label{fig:hg_v_resnet}
\end{figure}

\custompar{\modelnamenospace.}
In  Figure~\ref{fig:hg_v_resnet} we plot the results of a similar study for our anchor-free \modelname model, this time cropping only to groundtruth but evaluating even more mask-head variants.
And again we see a similar trend ---
while  mAP on the seen classes depends a little bit on the specific mask-head
architecture, the effect is small ($38.8\rightarrow40.0$).
However, for the same settings, the mAP on unseen classes varies significantly ($25.0\rightarrow32.5$).
We also see here that depth plays a role: empirically, it is important to go significantly beyond 4 layers
to achieve the best performance.
From a classical perspective, this is counterintuitive given the 
over-parameterization of very deep mask-heads, but perhaps is not so 
surprising in light of recent ways of rethinking generalization for
deep learning~\cite{zhang2016understanding,zhang2019identity}.

However, depth is not the only factor %
that drives generalization;
Among the alternatives, the Hourglass mask-heads provide
the best generalization performance to unseen
classes for both \modelname and \modelnamemaskrcnnnospace.
And this is fortunate since %
it is also the most memory-efficient mask-head
due to successive downsampling layers. %

Finally, in Table~\ref{tab:res_backbone}, we show that 
our findings
are not tied to our  choice of Hourglass backbone. While
comparing mask-heads when using ResNet-FPN and Hourglass backbones,
we observe that performance on unseen classes
is lower with ResNet backbones, but that mask-head architecture still
strongly impacts generalization to unseen classes.

\begin{table}
\tablefontsize
\centering
\begin{tabular}{lrrcrr}
\toprule
{\bf Mask-Head} & \multicolumn{2}{c}{\bf ResNet-101-FPN} & &\multicolumn{2}{c}{\bf Hourglass-104}\\
\cmidrule{2-3} \cmidrule{5-6}
& {\bf Box} & {\bf Mask} && {\bf Box} & {\bf Mask} \\
\midrule
ResNet-4 & 32.6 & 22.6 &&  39.7 & 26.6\\
Hourglass-10 & 32.2 & 24.8  && 39.9 & 29.4\\
Hourglass-20\ \ \ \  & 32.5 & 26.7 && 39.7 & 32.5\\
\bottomrule
\end{tabular}
\caption{Effect of \modelname backbones on the performance of various mask-heads.
Note that the box mAP is relatively unchanged
as we train with all boxes.
We train with the \voctrain~ setting and report mask mAP.}
\label{tab:res_backbone}
\end{table}

\begin{table}
\tablefontsize
\centering
\begin{tabular}{lrrr}
\toprule
{\bf Mask-Head} &\multicolumn{3}{c}{\bf mIOU} \\
\cmidrule{2-4} 
& {\bf Overall} & \ \ {\bf VOC} & {\bf Non-VOC} \\
\midrule
ResNet-4  & 67.0 & 78.6 & 62.1 \\
Hourglass-20  & 78.6 & 81.0 & 77.8 \\
Hourglass-52  & 78.9 & 81.1 & 79.2 \\
\bottomrule
\end{tabular}
\caption{mIOU of \modelname trained without any detection losses
under the \voctrain~setting.
Because we cannot compute mask mAP without a detector, we report mIOU
computed over the full validation dataset and over VOC/non-VOC class splits.
Hourglass mask-heads continue to show strong mask generalization on non-VOC classes,
even when they are not coupled with a detector.
}\vspace{-2mm}
\label{tab:iou_nodet}
\end{table}

\custompar{Strong mask generalization without a detector.}
\label{sec:detectionfree} 
To further make the point that the detection architecture
does not play a critical role in our story, we consider a ``detection-free''  incarnation of our model family, in which we do not even require the model to produce detections.  In this most basic of settings, we assume that a groundtruth box for each instance is provided as input and the task is to simply produce the correct segmentation mask.  For this setting, 
we use the \modelname architecture with an Hourglass backbone,
cropping to each groundtruth box and passing the result to the mask-head.  Since detection is no longer a task of interest, we drop all detection related losses and train only with sigmoid cross entropy loss for the masks.  We also evaluate using the mean IOU metric instead of mask mAP.

Table \ref{tab:iou_nodet} shows the results of this experiment
using 3 different mask-head architectures.  We observe that all architectures have similar performance on the seen categories ($\sim$2.5\% spread) whereas on unseen categories, the best mask-head (Hourglass-52) outperforms the worst (ResNet-4) by $>$16\%.
This confirms the strong mask generalization effect occurs in the detection-free setting and together with our results
for \modelname and \modelnamemaskrcnn provide strong evidence that we would find similar effects using other detection
architectures.

\section{A closer look at mask-head architectures}
\label{sec:ablations}
\label{sec:mask_head_arch_ablations}
Having established that mask-head architecture significantly
affects strong mask generalization in crop-then-segment models, 
we now study the mask-heads in detail and identify
the components that are most crucial.
All ablations
in this section  are done on \modelname and
use the \voctrain~ setting for training
and evaluation.

\begin{table}
\tablefontsize
\centering
\begin{tabular}{llrrr}
\toprule

{\bf Mask-Head} & {\bf Variant} & \multicolumn{3}{c}{\bf Mask mAP} \\
\cmidrule{3-5} 
&& {\bf Overall} & \ \ {\bf VOC} & {\bf Non-VOC} \\
\midrule
ResNet-20 & Default & 31.4 & 39.1 & 28.8 \\
\midrule
\multirow[t]{3}{*}{Hourglass-20} & Default &  34.1 & 39.8 & 32.2 \\
& No LRS & 33.6 & 39.2 & 31.7 \\
& No ED & 31.7 & 39.1 & 29.2 \\
\bottomrule
\end{tabular}
\caption{%
Isolating what makes Hourglass architectures achieve strong mask generalization.
No LRS = No long range
skip connections. No ED = No encoder-decoder structure, i.e, no downsampling
or upsampling layers.
}\vspace{-1mm}
\label{tab:resnet_v_hourglass}
\end{table}

\begin{table}[t!]
\centering
\tablefontsize
\begin{tabular}{crrr}
\toprule
\multirowcell{2}{\bf \# Dilated conv layers \\
{\bf replaced}} &  \multicolumn{3}{c}{\bf Mask mAP} \\
\cmidrule{2-4} 
& {\bf Overall} & \ \ {\bf VOC} & {\bf non-VOC} \\
\midrule
0 &32.2 &39.4 &29.9 \\
10 & 32.7 &39.1 &30.6 \\
20 & 32.8 & 39.3 &30.7 \\
\bottomrule
\end{tabular}
\caption{
Replacing regular convolutional layers with dilated convolutions (rate=2)
in a ResNet-20 mask-head to isolate the effect of receptive field. 
}
\label{tab:dilate_resnet20}
\end{table}

\begin{table}
\tablefontsize
\centering
\begin{tabular}{llrrr}
\toprule
{\bf Backbone} & {\bf Mask-Head} &  \multicolumn{3}{c}{\bf mIOU} \\
\cmidrule{3-5} 
&& {\bf Overall} & \ \ {\bf VOC} & {\bf Non-VOC} \\
\midrule
HG-52 & HG-52 & 78.4 & 80.4 & 77.8 \\
HG-104 & ResNet-4 & 71.4 & 79.2 & 68.8 \\
\bottomrule
\end{tabular}
\caption{
Can we reproduce strong mask generalization
by adding an hourglass network to the shared backbone instead of using it in the per-proposal mask-head?
We compare two networks of similar depth where the first network has a deeper mask-head.
For fair comparison, we use groundtruth boxes as input 
at evaluation time and report mIOU.
}\vspace{-1mm}
\label{tab:shallower_fe}
\end{table}

\custompar{What makes Hourglass mask-heads so good?}
We first address the question of which
architectural elements are most responsible for the superior generalization
of Hourglass networks.
To investigate, we focus on 20-layer Hourglass and ResNet basic
mask-heads.
The Hourglass architecture differs in two main ways from ResNet, having
a) an encoder-decoder
structure in which the encoder downsamples the input and the decoder
upsamples the result of the encoder, and
b) long range skip connections connecting
feature maps of the same size in the encoder and decoder.

To understand each difference in isolation,
we explore the effect of (a) replacing the downsampling/upsampling
layers with layers that do not change the feature map resolution, and 
(b) severing long range skip connections. 

Table~\ref{tab:resnet_v_hourglass} shows the corresponding ablation results.
We see that removing the long range skip connections (No LRS) has
a small negative impact on the performance.
More importantly, we find that the majority of the gap between ResNet and Hourglass is closed
by getting rid of the encoder-decoder structure in the Hourglass
mask-head (No ED).  Given these results, we conclude that this style of downsampling followed by upsampling 
likely captures particularly appropriate inductive biases for segmentation.

\custompar{What's so special about the mask-head?}
Next, given that an hourglass mask-head offers generalization advantages, we ask:
could we reproduce these advantages by adding an hourglass network to the shared backbone instead of using it
in the per-proposal mask-head? In other words, what is so special about the mask-head?
Here we show that the answer is negative and that it is indeed the mask-head's architecture
that impacts strong mask generalization.

Consider an  HG-104 network which is a stack of two 
hourglass modules each with 52 layers.  We compare (a) a model
where all 104 layers are situated in the backbone and we use a simple ResNet-4 mask-head
versus (b) a model with an HG-52 backbone and HG-52 mask-head.
In both cases, inputs undergo roughly 100 layers contained within two hourglass modules but in the second case,
the 52 layer mask-head is applied on a per-proposal basis.

Since using 52 layers in the backbone in general yields
inferior detection quality compared to the 
104 layer backbone, we use groundtruth boxes as input so that both models are on equal footing
and we evaluate mIOU. 

Our finding (Table~\ref{tab:shallower_fe}) is that despite having slightly fewer total layers, our model with the 52 layer mask-head
outperforms the model with the 4 layer mask-head by 9\% mIOU
on unseen classes
(both models have similar performance on seen classes).
More generally, this supports our hypothesis that within the entire architecture 
the mask-head plays a disproportionately significant role with respect to generalization
to unseen classes.

\begin{table*}[t!]
\setlength{\tabcolsep}{4pt}
\tablefontsize
\centering
\begin{tabular}{lrcrrrrrrcrrrrrrr}

\toprule
{\bf Model} & {\bf {b-box.}} && \multicolumn{6}{c}{\bf VOC $\rightarrow$ Non-VOC (mask) } & & 
\multicolumn{6}{c}{\bf Non-VOC $\rightarrow$ VOC (mask) } 
& {\bf ms./im}\\
\cmidrule{2-2}  \cmidrule{4-9} \cmidrule{11-16}

& $AP$ && $AP$ & $AP_{50}$ & $AP_{75}$ & $AP_{S}$ & $AP_{M}$ & $AP_{L}$ & \hspace{5pt} & 
$AP$ & $AP_{50}$ & $AP_{75}$ & $AP_{S}$ & $AP_{M}$ & $AP_{L}$ \\
\midrule
Mask R-CNN \cite{segment_everything} & 38.6 && 18.5 & 24.8 & 18.1 & 11.3 & 23.4 & 21.7 & & 24.7 & 43.5 & 24.9 & 11.4
& 25.7 & 35.1 & 56\\
Mask GrabCut \cite{segment_everything} & 38.6 && 
19.7 & 39.7 & 17.0 & 6.4 & 21.2 & 35.8 &\hspace{10pt}& 19.6 & 46.1 & 14.3 &  5.1 &  16.0 & 32.4 & ---\\
Mask$^{X}$ R-CNN \cite{segment_everything} &  38.6 & &23.8 & 42.9 & 23.5 & 12.7 & 28.1 & 33.5  && 29.5 & 52.4 & 29.7 & 13.4 & 30.2 & 41.0 & ---\\
ShapeMask \cite{shapemask} & 45.4 & &
33.2 & 53.1 & 35.0 & 18.3 & 40.2 & 43.3 && 35.7 & 60.3 & 36.6&  18.3 & 40.5 & 47.3 & 224\\
CPMask \cite{cpnet} & 41.5 & &
34.0 & 53.7 & 36.5 & 18.5 & 38.9& 47.4 && 36.8 & 60.5 & 38.6 & 17.6 & 37.1 & 51.5
& ---\\
\modelname(ours) & 44.5 & &
35.5  &  54.6  &  38.2 & 19.4 & 40.3 &  50.6 & &
 39.1 &  62.6 & 41.9 & 17.6 &  38.7 &  54.0 & 232\\
\modelnamemaskrcnn (ours) & 48.6 & &{\bf 38.7} & {\bf 62.5} & {\bf 41.0} &{\bf 22.3} &
{\bf 43.0} &{\bf 55.9}
&&
{\bf 41.0} & {\bf 68.2} & {\bf 43.1} & {\bf 22.0} & {\bf 40.0} & {\bf 55.9}
& 170
\\

\bottomrule
\end{tabular}
\caption{
Partially supervised performance of \modelname (CenterNet based) and \modelnamemaskrcnn (Mask R-CNN based) compared to other models.
We measure mask mAP on the \texttt{coco-val2017} set.
The top row with label A $\rightarrow$ B indicates that
we train on masks from set A and evaluate our masks on set B.
Bounding box (b-box.) AP is an average over all classes.
We use report inference time as milliseconds / image (ms./im)
on a V100 GPU and compare with Detectron2 \cite{detectron2} 
and ShapeMask\cite{shapemask}. 
CPMask\cite{cpnet}, Mask$^X$\cite{segment_everything} R-CNN have not reported inference time.
}\vspace{-2mm}
\label{tab:highres}
\end{table*}

\custompar{Is it sufficient to have a large receptive field?}
Finally, given that depth and encoder/decoder structures do so well,
it seems natural to conjecture that increased receptive field in these architectures may play a 
significant role.

To evaluate this hypothesis, we explore two additional families of mask-heads:
(a) we replace the vanilla convolutions in a ResNet mask-head with dilated
convolutions (w/ rate 2), which has the effect of expanding receptive field without changing
the depth or number of parameters, and (b) we use fully connected (MLP) mask-heads which have
full receptive field. See Table~\ref{tab:dilate_resnet20} and Appendix~\ref{sec:fcn} for dilated and FC results, respectively.

Our experiments using both families of models
show, first, that none of these models are able to 
reach the performance of Hourglass mask-heads, so there must be further factors at play beyond
receptive field.  On the other hand, 
growing the receptive field early seems to benefit generalization to some extent (e.g., shallow FC mask-heads outperform
shallow convolutional mask-heads).  

And this raises an interesting question which we leave for further study:
what about receptive field would help
unseen classes without simultaneously helping seen classes?
Here we offer one conjecture based on
our Mask R-CNN finding (Section~\ref{sec:importance_gt}), 
that it is important to train using groundtruth boxes instead of proposals.
A groundtruth box, when tight on an instance, acts as a cue, indicating the
object that is meant to be segmented.  When trained on noisy proposals, we conjecture
that Mask R-CNN tries to memorize the types
of foreground classes seen at training time and thus struggles to generalize to unseen classes.
With a precise cue, however, perhaps 
the model learns to compare interior pixels to boundary pixels
to make this decision, a strategy that is more generalizable across categories
and requires a large enough receptive field so that boundary pixels can interact with interior pixels.

\vspace{-2mm}
\section{Comparison with the state-of-the-art}\label{sec:sota}
We now train models at higher resolutions that previous
sections with \modelname being trained at $1024\times 1024$
and \modelnamemaskrcnn at $1280 \times 1280$.
\modelname uses an Hourglass-104 backbone and an Hourglass-100
mask-head, whereas \modelnamemaskrcnn uses a SpineNet-143 \cite{spinenet}
backbone and an Hourglass-52 mask-head.
With these settings, \modelname and \modelnamemaskrcnn
beat previous state-of-the-art approaches as 
seen in Table \ref{tab:highres}.
\modelnamemaskrcnn produces our best result
which exceeds CPMask\cite{cpnet} on VOC to Non-VOC transfer by $4.7\%$ and
Non-VOC to VOC transfer by $4.2\%$.
Compared to  prior approaches, our method is end-to-end trainable
and does not require auxiliary losses or specialty modules. %
Although \modelname surpasses the state of the art by itself, we 
 show in the Appendix (Section~\ref{sec:distiallation}) that we 
 can do even better using distillation based training
(achieving a Non-VOC mAP of $40.4\%$ on the same problem).

\vspace{-2mm}
\section{Conclusions}\label{sec:conclusion}
In this work, we have identified and studied the surprising
extent to which the mask-head architecture impacts 
generalization to unseen categories as well as the connection between this effect
and the protocol of cropping to only groundtruth boxes at training time.  
Through extensive experiments, we demonstrated the generality
of this effect across detection methodologies and backbone networks. 
And by exploiting this strong mask generalization
effect, we established a
new state of the art on this problem by a significant margin using a conceptually simple model.

While we have taken initial steps in understanding strong mask generalization, 
how to better understand the inductive biases encoded within mask-head architectures and
how to explain our
results theoretically remain important directions.
Along these lines, we leave readers with pointers to two  papers which have noted similar 
empirical phenomena where certain architectures %
generalize effectively to data outside of 
the training distribution.
The Deep Image Priors work~\cite{ulyanov2018deep} similarly 
observed that Hourglass-style networks seem to automatically capture image level statistics 
in a natural way without being trained on data. \cite{zhang2019identity} showed that sufficiently deep networks unlock a certain strong generalization behavior.  
We conjecture that there may be a common denominator at play and that exploring
these synergies further would be a fruitful area of further research potentially yielding insights useful beyond segmentation.

\custompar{Acknowledgements}
We would like to thank David Ross for thoughtful feedback and Pengchong Jin, Abdullah Rashwan and Xianzhi Du for helping with Mask R-CNN code.

{\small
\bibliographystyle{style/ieee_fullname}
\bibliography{main}
}
\clearpage

\begin{appendices}

\section{Additional Ablations}

\subsection{Fully supervised performance}

We report fully supervised performance of \modelname and \modelnamemaskrcnn
in Table~\ref{tab:full_super}.

\begin{figure*}[ht!]
    \centering
    \includegraphics[width=0.9\textwidth]{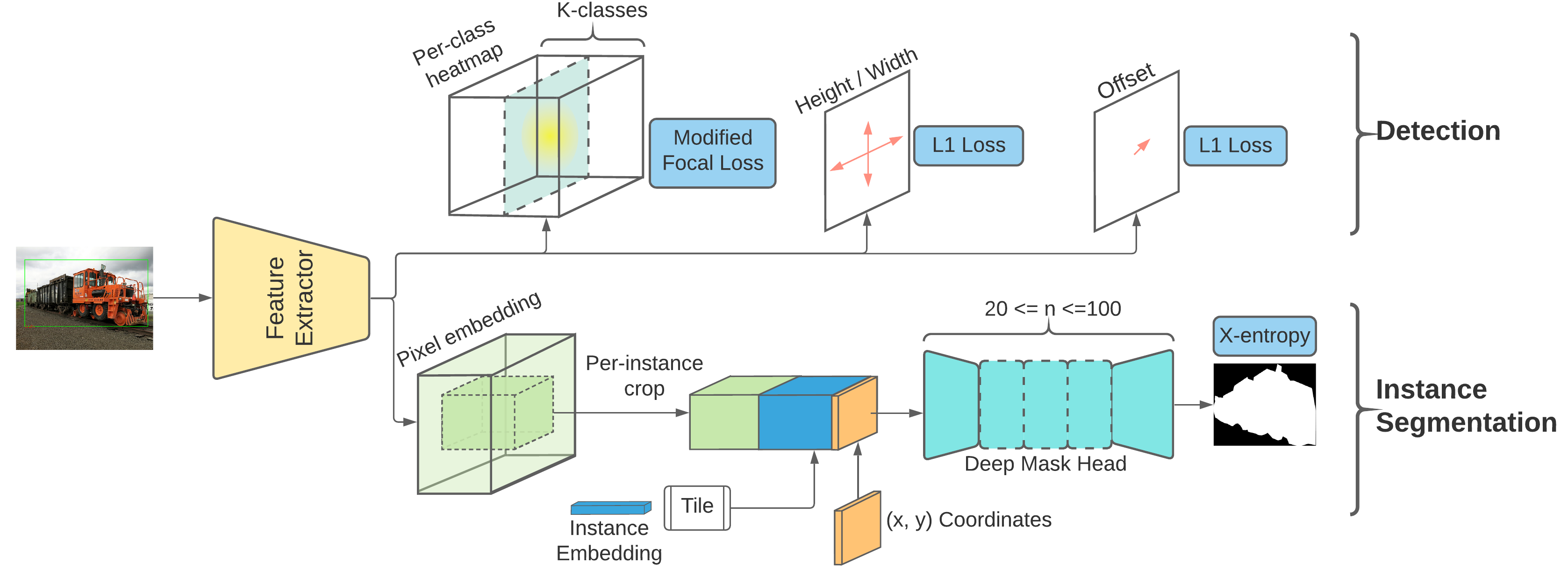}
    \caption{Schematic of the \modelname architecture. The top-half is kept 
    identical to CenterNet \cite{centernet} and the bottom-half uses an RoI
    crop followed by a deep mask head. In our experiments, it was crucial
    to train the mask head with only groundtruth boxes.}
    \label{fig:diagram}
\end{figure*}
\begin{table}[h!]
{
\tablefontsize
\centering
\begin{tabular}{llrrrr}
\toprule
{\bf Model} & {\bf Backbone}&  $AP$ & $AP_{S}$
& $AP_{M}$ & $AP_{L}$ \\
\midrule
ShapeMask  & RF101 & 37.4 &  16.1 & 40.1 & 53.8 \\
ShapeMask & RNF101 & 40.0 & 18.3 & 43.0 & 57.1 \\
CPMask~  & RF101 &
39.2  & 22.2 & 41.8 & 50.1 \\
\modelname  & HG104 & 39.4 & 20.5 & 41.9 & 54.0 \\
\modelnamemaskrcnn & SN143 & 42.8 & 24.3 & 46.0 & 60.5 \\
\bottomrule
\end{tabular}
\caption{ Fully supervised instance segmentation performance on COCO \texttt{test-dev2017}. Backbones include
RF=ResNet-FPN, 
RNF=ResNet-NAS-FPN, 
HG=Hourglass,
SN=SpineNet.
\modelname is trained at $1024\times1024$ resolution with an HG-100 mask-head
and \modelnamemaskrcnn is trained at $1280\times1280$ resolution with HG-52 mask-head.
Mask heads are explored in detail in Section \ref{sec:going_deeper_mask_heads}.
We report mAP of \texttt{coco-testdev2017}.
}

\label{tab:full_super}
}
\end{table}

\subsection{\modelnamenospace}
\subsubsection{Effect of instance and coordinate embedding}
\label{sec:emebdding_ablations}
Table \ref{tab:embedding} shows the effects of coordinate embedding
and instance embedding on ResNet and Hourglass mask heads.
We notice that the additional embeddings do not make a significant difference
to the Hourglass model, but coordinate embedding is required for 
the ResNet based mask heads to converge.  For uniformity, we have thus used both components in all
\modelname variants.
\begin{table}[h]
\centering
\begin{tabular}{lccrrr}
\toprule
{\bf Mask Head} & {\bf C} & {\bf I} & \multicolumn{3}{c}{\bf Mask mAP} \\
\cmidrule{4-6}
 &&& {\bf \small Overall} & {\bf\ \small  VOC} & {\bf\small Non-VOC} \\
\midrule
\multirow[t]{4}{*}{ResNet-20} & & &-- & --  & --\\
&& \ding{51} & -- & -- & -- \\
& \ding{51} & &30.9 & 39.1 & 28.2 \\
& \ding{51} & \ding{51} & 31.4 & 39.1 & 28.8 \\
\midrule
\multirow[t]{4}{*}{HG-20}
& & & 34.1 & 39.8 & 32.2 \\
& & \ding{51} & 34.5 & 39.9 & 32.7\\
& \ding{51} & & 33.6 & 39.9 & 31.5 \\
& \ding{51} & \ding{51} &
34.3 & 39.8 & 32.5 \\
\bottomrule
\end{tabular}
\caption{ Effect of Coordinate Embedding (C) and Instance Embedding (I) on the generalization ability of \modelname on unseen classes. A `--' indicates that the model failed to converge. All models are trained with masks only from VOC classes at an input
image resolution of $512 \times 512$.
Performance is reported with the \voctrain~ setup.}
\label{tab:embedding}
\end{table}
\subsubsection{Effect of using fully connected layers}
\label{sec:fcn}
See Table~\ref{tab:fcn} for experiments with fully connected layers. 
We used Glorot normal initialization~\cite{glorot} the mask-head weights.
Based on these results, we see that the fully connected mask-head models, which have 
full receptive field with respect to the input tensor, do not offer competitive
performance compared to the HG-based mask-heads.  However, early large receptive fields
may still be beneficial to some extent as these fully connected mask-heads do outperform
our shallowest convolution-only mask-heads (e.g. Resnet-4).
\begin{table}
\centering
\begin{tabular}{crrr}
\toprule
{\bf FCN layers} & \multicolumn{3}{c}{\bf Mask mAP} \\
\cmidrule{2-4}
 & {\bf Overall} & {\bf\ \  VOC} & {\bf Non-VOC} \\
\midrule
2 & 29.1 & 38.4 & 26.0\\
4 & 30.5 & 37.5 & 28.2\\
\bottomrule
\end{tabular}
\caption{ Effect of using fully connected layers as mask-heads
on \modelname.
Performance is reported with the \voctrain~ setup.
For easy reference, the VOC/non-VOC mask mAP values for
Resnet-4 and HG-52 mask-heads are 39.7/26.6 and 39.8/32.5 respectively.
}
\label{tab:fcn}
\end{table}

\subsection{Mask R-CNN}
Table~\ref{tab:maskrcnn_supervised} shows the impact of using groundtruth boxes 
(instead using proposals, which is the standard approach)
for training the mask-head of a fully supervised Mask R-CNN model on COCO. First
we see that using
a class-agnostic mask head results in a slightly lower mask mAP compared to the standard
class-specific mask-head.  
Training with groundtruth boxes instead of proposals does not further impact the performance of the class
agnostic mask head significantly.

\begin{table}
\centering
\begin{tabular}{lrrr}
\toprule
{\bf Variant} & {\bf Mask mAP} \\
\midrule
Class-specific (Proposals + GT) & 37.2 \\
Class-agnostic (Proposals + GT) & 36.7\\
Class-agnostic (GT only) & 36.4 \\
\bottomrule
\end{tabular}
\caption{
Fully supervised mask mAP of Mask-RCNN variants with a ResNet-50-FPN backbone.
}
\label{tab:maskrcnn_supervised}
\end{table}

\section{Using \modelname just for its masks for two-stage training}\label{sec:deepmac_just_masks}
In this section we show that it is detection quality rather than mask quality which is now
the bottleneck to achieving even better performance on the partially
supervised COCO task, at least with respect to the mAP metric.
With this insight, we use Deep-MAC to label instance masks on 
classes where they do not already exist and train a model with better
detection performance on the resulting data.

\subsection{Limited headroom on COCO mask quality.}
\label{sec:limited_headroom}
If, as with the detector-free model from Section~\ref{sec:detectionfree}, we run the mIOU evaluation on the Deep-MAC model cropping to groundtruth boxes, we obtain 81.4\% which is slightly better than the detection-free model.  To put this number in perspective, \cite{gupta2019lvis} showed that COCO groundtruth masks achieve 83\%-87\% mIOU when compared to expert labels.  Thus our finding suggests that remaining headroom on improving segmentation quality is quite limited (we are likely close to a saturation point).
Note that this is not to say that our models have reached human level performance, since COCO annotation quality is known to be lower 
that some more recent datasets (e.g., LVIS~\cite{gupta2019lvis}).
However, our finding does suggest that future improvements on the partially supervised task on COCO as measured by mean AP will be much easier to come by via improvements to detection quality as opposed to segmentation quality.

\begin{table}[t!]
\centering
\tablefontsize
\begin{tabular}{llrrr}
\toprule
{\bf Model} & {\bf B.B.} &  \multicolumn{3}{c}{\bf Mask mAP} \\
\cmidrule{3-5} 
&& {\bf Overall} & \ \ {\bf VOC} & {\bf non-VOC} \\
\midrule
\modelname[R4] & HG104 & 37.8 & 42.2 & 36.3 \\
Mask R-CNN  & RF50  & 36.1 & 40.2 & 34.7 \\
Mask R-CNN & SN143  & 41.9 & 46.4 & {\bf 40.4} \\
\bottomrule
\end{tabular}
\caption{
Using \modelname generated pseudo labels to train other models.
\modelname is trained as described in Table~\ref{tab:full_super} on
pseudo labels
and evaluated on the \texttt{coco-val2017} set.
Other models are trained with their default settings.
Backbones(B.B.) include
HG=Hourglass, RF=ResNet-FPN, SN=SpineNet. R4=ResNet-4 mask-heads.
For reference, the ``teacher'' \modelname model achieves a non-VOC
mAP of ~35.5\% (c.f. Table~\ref{tab:highres}).
}
\label{tab:pseudo_gt}
\end{table}

\subsection{Two stage (self-distillation style) training for improved mAP or cheaper models.}
\label{sec:distiallation}

To illustrate, we use Deep-MAC just for its masks (and not its boxes), first segmenting unseen categories and then training a stronger detection model (Mask R-CNN with SpineNet~\cite{spinenet}, which reaches 48.6\% box AP compared to 
\modelname which reaches 44.1\% box AP) in fully supervised mode on these pseudo labels.
Table~\ref{tab:pseudo_gt} (last row) shows the result of this experiment — specifically, Mask R-CNN with SpineNet is able to leverage the pseudo labels
to get to a 40.4\% non-VOC mask mAP, which is significantly higher than the original model that generated the pseudo labels. Thus  improving box detection quality leads to a significantly increased final non-VOC mAP which is not upper bounded by the non-VOC mAP of \modelname itself.
This is also the highest performance ever reported on the partially supervised task by a margin of $6.4\%$ (but only by virtue of better detection and without improving generalization to novel classes).

Our recommendation, consequently, is that the community should focus on harder tasks either by training with even fewer mask annotations, or 
evaluate partially supervised performance on LVIS~\cite{gupta2019lvis}  which has more classes and higher quality masks.
As an initial step, we train \modelname on COCO
masks from all 80 categories and evaluate mIOU on LVIS masks (from the \texttt{v1-val} set) cropping to
LVIS groundtruth boxes. Here our models using ResNet-4 and HG-100 mask-heads achieve 70.3\% and 79.9\% mIOU respectively,
showing that architecture continues to matter for strong mask generalization even on LVIS.  Comparing to~\cite{gupta2019lvis}
who report 90-92\% mIOU dataset-to-expert agreement, we also see that 
there is still a gap between \modelname and human performance (but this is likely
at least in part due to COCO's lower quality masks).

Another application of  two stage training  is to train a cheaper instance segmentation model on masks produced by Deep-MAC.  The first two rows of Table~\ref{tab:pseudo_gt} demonstrate results using a cheaper Mask R-CNN model or Deep-MAC model with a shallower (4 layer) mask-head.  
This experiment is particularly interesting in the case of the student Deep-MAC model with the shallow head since in this two stage setting, the student trains as if it were being fully supervised. 
According to our
findings in Section~\ref{sec:going_deeper_mask_heads}, we should therefore expect it to achieve 
the same performance as Deep-MAC with the heavier mask-head (which it does, and even exceeds).  
Thus for COCO categories we are able to leverage the strong mask generalization properties of the heavier mask-head while retaining the computational benefits of the cheaper mask-head. 
When running at $1024 \times 1024$ resolution on a V100 GPU,
\modelname with an HG-100 mask-head takes 232 ms per image, whereas the cheaper student model with a ResNet-4 mask-head %
is faster (201 ms per image).
Notably, this cheaper student model
is on par with ShapeMask~\cite{shapemask} in terms of speed (204 ms) while achieving a
2.1 \% improvement on non-VOC mAP.\footnote{Inference speed is not reported in \
CPMask~\cite{cpnet}}

\section{Generalizing from a single class}
\label{sec:single_class}

\begin{figure*}[t]
\centering
 \begin{subfigure}[t]{0.6\linewidth}
     \centering
     \includegraphics[width=0.9\linewidth]{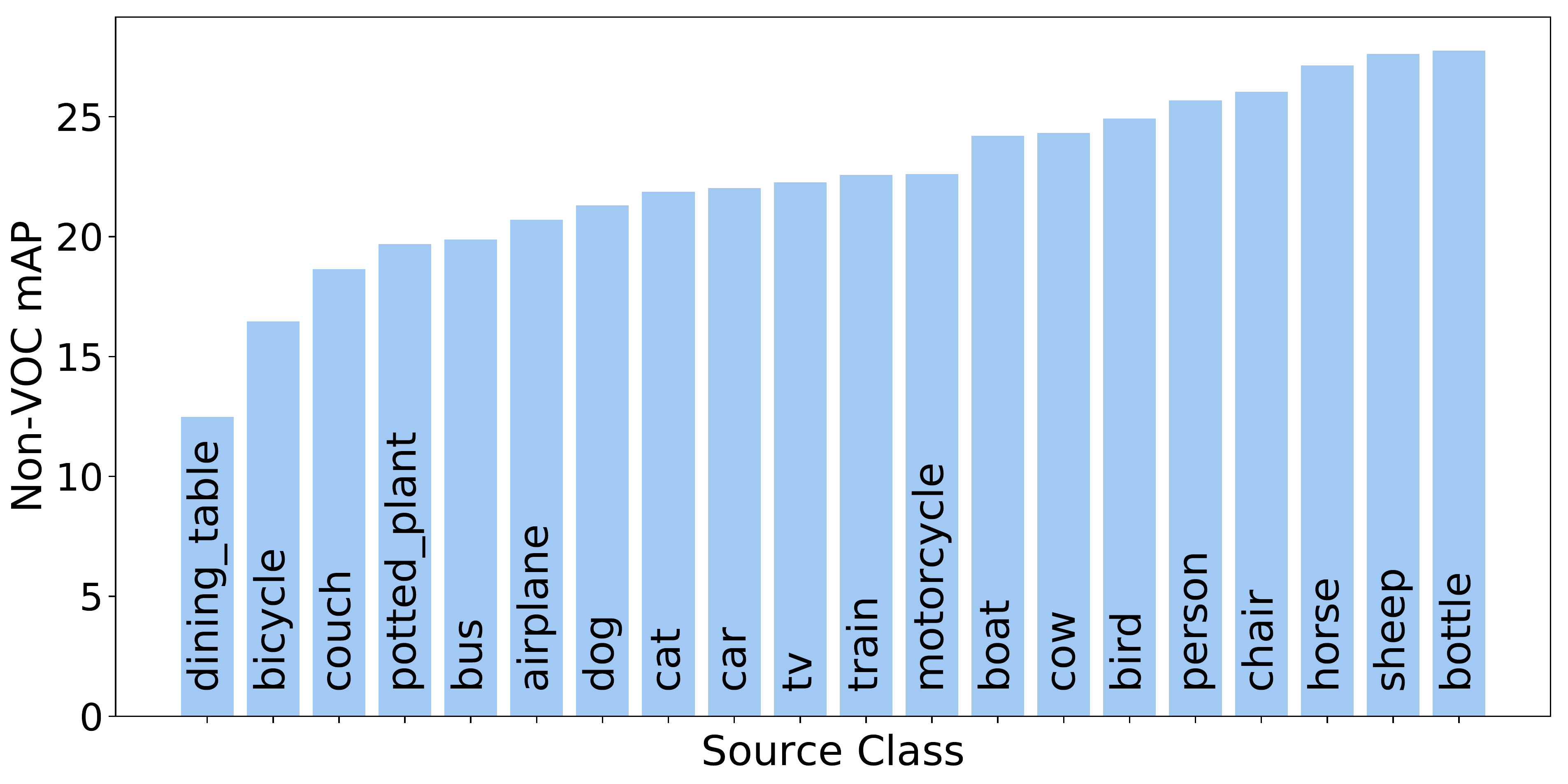}
     \vspace{-7pt}
     \caption{}
     \label{fig:bar}
 \end{subfigure}

 \begin{subfigure}[t]{0.6\linewidth}
     \centering
     \includegraphics[width=1.0\linewidth]{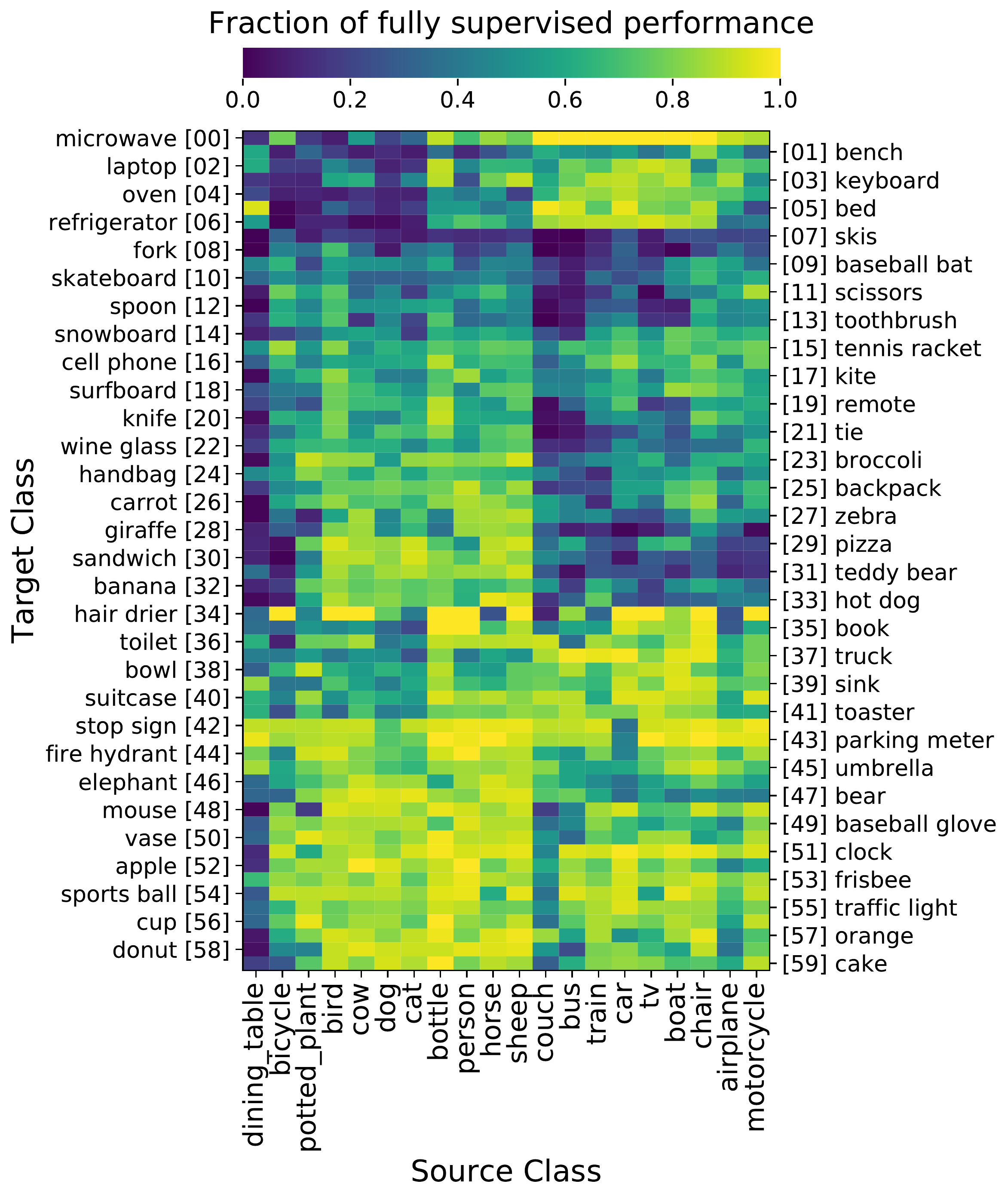}
     \vspace{-20pt}
     \caption{}
     \label{fig:heatmap}
 \end{subfigure}
\caption{
(\subref{fig:bar}) Mask mAP on Non-VOC classes when training with
masks from only a single source class from the VOC set;
(\subref{fig:heatmap}) Performance on specific (Non-VOC)
target classes when training with masks from only a single class.
We visualize  performance relative to full supervision. 
}\vspace{-1mm}
\label{fig:single_class} 
\end{figure*}

In the majority of our experiments, we assumed the standard setup of ``train-on-VOC, test-on-non-VOC''.  In this section, 
we restrict further, training on a single ``source'' class at a time, in order to better understand when 
\modelname can be expected to strongly generalize to a novel class.
In Figure \ref{fig:bar} we plot results from %
this experiment, training on each of the VOC categories with $512\times 512$ resolution inputs and an Hourglass-52 mask head.
We observe that while some classes lead to strong performance, there is
high variance depending on the source category
(ranging from 12.5\% mAP to 27.8\% mAP).  Notably, a single class can achieve strong results --- as one datapoint, training only on the chair category with higher resolution $1024\times 1024$ inputs
yields a non-VOC mask mAP of 31.5, which is competitive with previous high-performance methods (e.g., ShapeMask~\cite{shapemask}) when trained on all VOC categories.

In some cases it is easy to guess why a class might be a poor source --- on the worst classes, we see that the quality of groundtruth masks is uneven in COCO.  For example, labelers were not consistent about excluding objects that were on but not part of a dining table (see
Section~\ref{sec:annotation_quality}).  

For more detailed insight, we ask how training on a specific source class might generalize to a specific new target class. For source-target pairs $(i, j)$, Figure~\ref{fig:heatmap} visualizes this relationship via the ratio between mAP on target class $j$ if we were to train on just the source class $i$ to mAP on target class $j$ if we were to train on all classes.  Here we cluster the rows and columns by similarity and truncate ratios to be at most 1.0 for visualization purposes.\footnote{It is worth noting in several cases (most notably, hair drier [34]) that it is better to train on other source classes than it is to include the target class annotations during training.}

Figure \ref{fig:heatmap}  illustrates that some classes (e.g., apple [52], umbrella [45], stop sign [42]) are universally easy transfer targets likely due to being visually salient,  having consistent appearance and  not typically co-occurring with other examples of their own class.
We also see that co-occurrence of source and target classes does not always lead to improved ratios (i.e. close to 1).  For example, training on car does not yield strong performance on stop signs [42] or parking meters [43] and training on person does not yield strong performance on bench [01] or baseball bat [09]. On the other hand, categories that are similar semantically seem to function similarly as source categories, and with a few exceptions, the source categories cluster naturally into two broad groups: man-made and natural objects. 

It remains an open question why a class might excel as a source class in general.  Intuitively one might think that person, car or chair categories might be the best because they have the most annotations and are visually diverse, but perhaps surprisingly, using the bottle category is the best.  This may be due to the fact that bottles tend to look alike and appear in groups, forcing the model to make non-local decisions about mask boundaries.  We leave exploration of this hypothesis for future work. %

\section{Example images on unknown classes}
\begin{figure*}
\centering
\begin{subfigure}[t]{0.40\linewidth}
\includegraphics[width=1\linewidth]{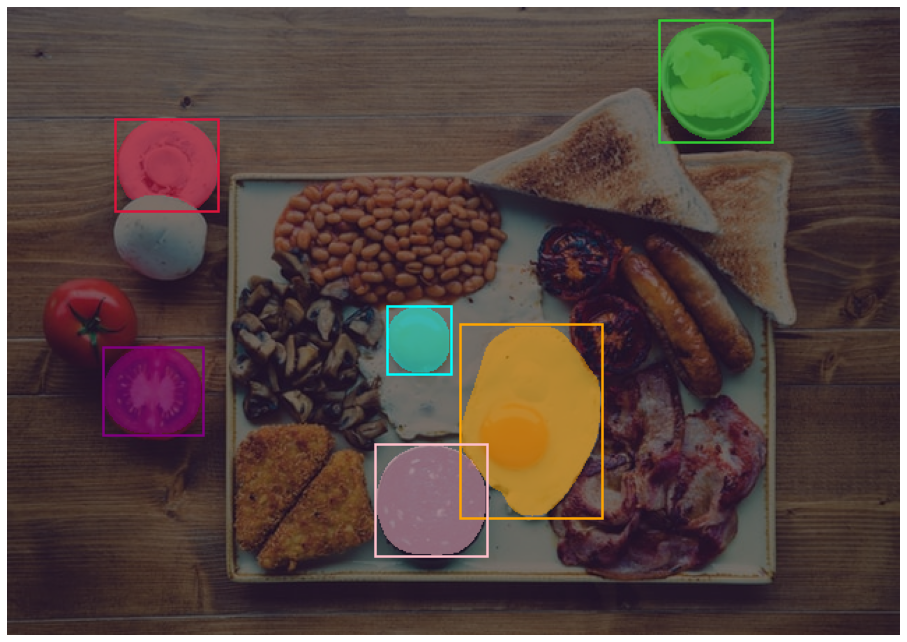}
\caption{Photo by Jonathan Farber on Unsplash.}
\end{subfigure}
\begin{subfigure}[t]{0.40\linewidth}
\includegraphics[width=1\linewidth]{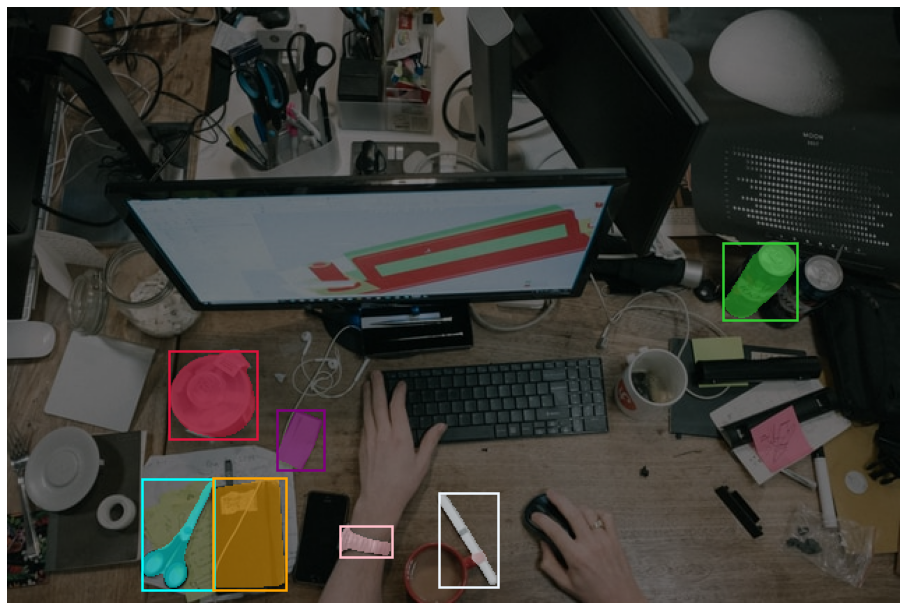}
\caption{Photo by Robert Bye on Unsplash.}
\end{subfigure}
\begin{subfigure}[t]{0.40\linewidth}
\includegraphics[width=1\linewidth]{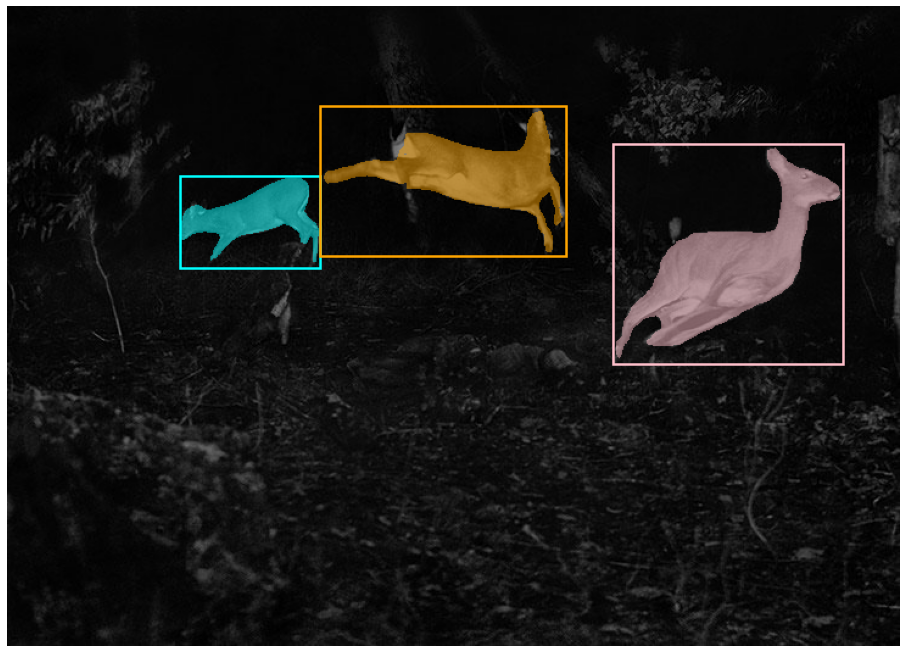}
\caption{Sample from the Snapshot Serengeti dataset.}
\end{subfigure}
\begin{subfigure}[t]{0.40\linewidth}
\includegraphics[trim=0 0 0 200,clip,width=1\linewidth]{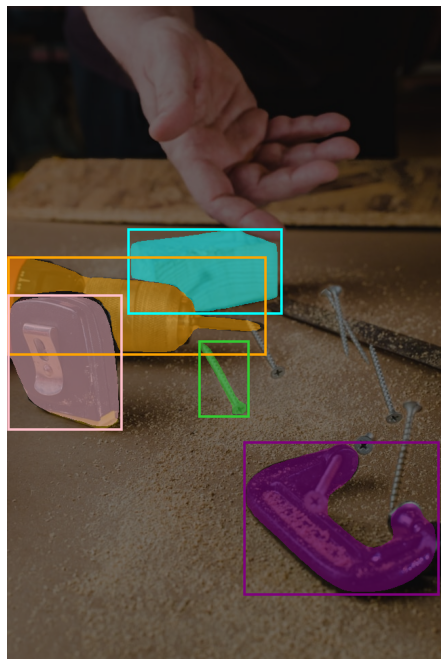}
\caption{Photo by Chris Briggs on Unsplash.}
\end{subfigure}
\begin{subfigure}[t]{0.30\linewidth}
\includegraphics[width=1\linewidth]{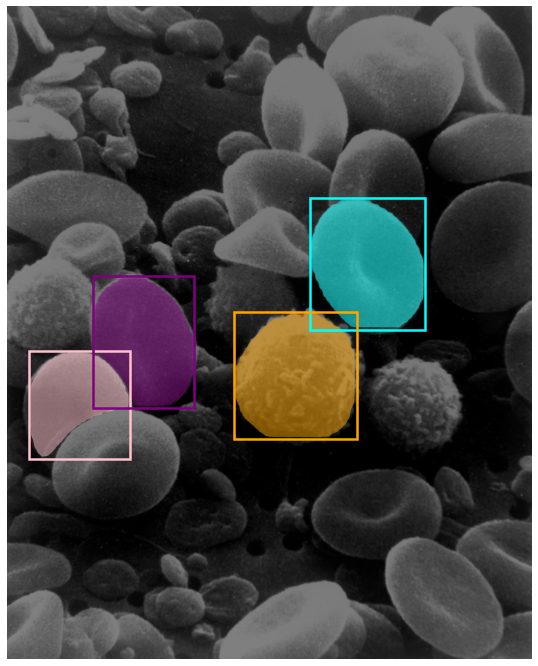}
\caption{SEM blood cells image from wikipedia.}
\end{subfigure}
\begin{subfigure}[t]{0.40\linewidth}
\includegraphics[width=1\linewidth]{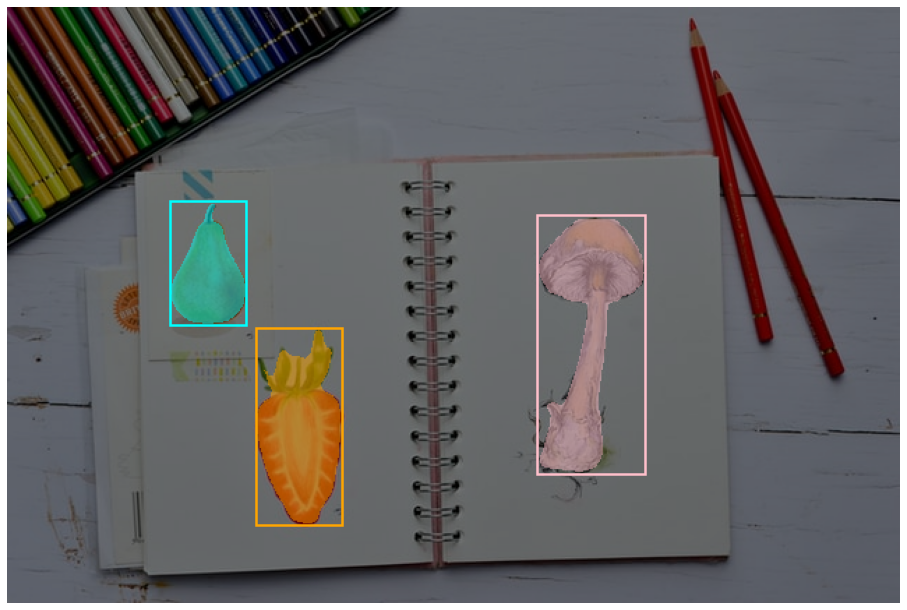}
\caption{Photo by Maggie Jaszowska on Unsplash.}
\end{subfigure}
\caption{Example outputs of \modelname with hand-drawn boxes on unknown classes.}
\label{fig:user_boxes}
\end{figure*}

See Figure~\ref{fig:user_boxes} for example outputs of \modelname.
We look at the output of our model with user-specified boxes around object 
categories that are not in the COCO dataset. We observe that \modelname generalizes
to multiple different domains like biological and camera trap images and 
does well even in cluttered settings.
For this experiment, we used a model trained with all COCO classes in
fully supervised mode.

\section{Looking at annotation quality}
\label{sec:annotation_quality}
\begin{figure*}
\centering
\begin{subfigure}[t]{0.93\linewidth}
\includegraphics[width=0.32\linewidth]{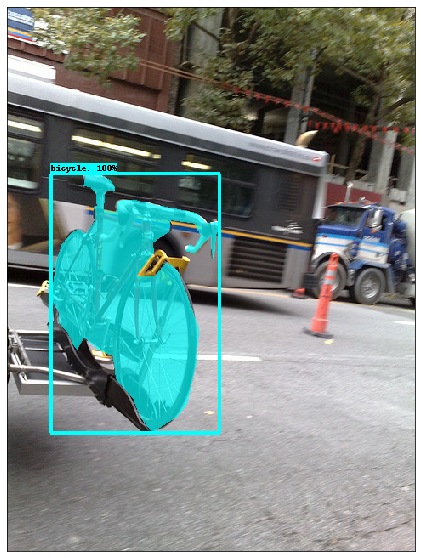}
\includegraphics[width=0.32\linewidth]{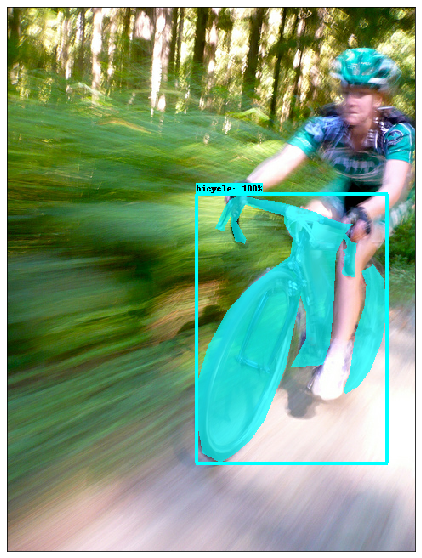}
\includegraphics[width=0.32\linewidth]{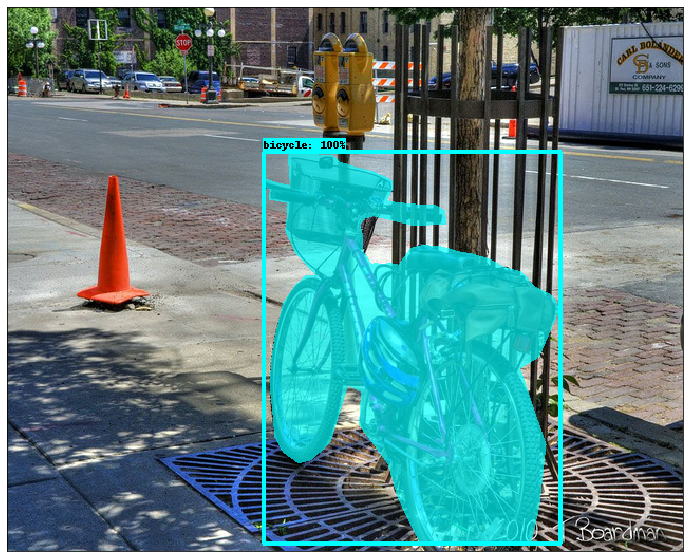}
\caption{Bicycle: The annotated masks don't capture the shape correctly,
and quite often label parts of the background interspersed with the bicycle
frame as foreground.}
\end{subfigure}

\begin{subfigure}[t]{0.93\linewidth}
\includegraphics[width=0.32\linewidth]{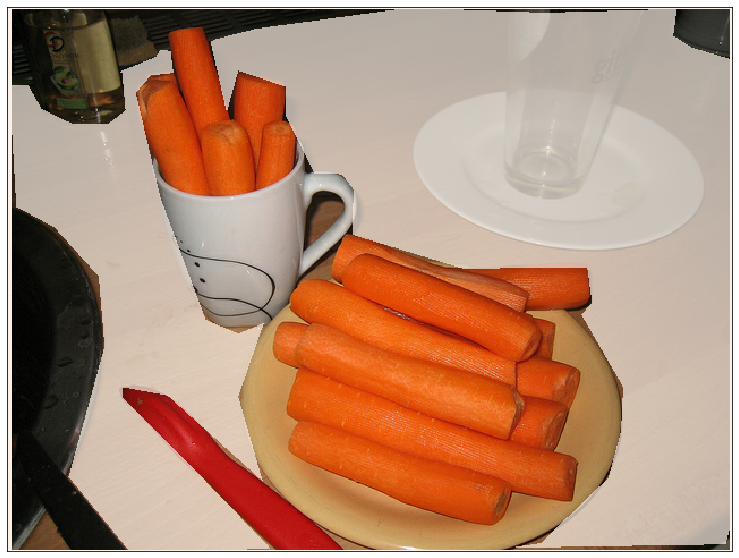}
\includegraphics[width=0.32\linewidth]{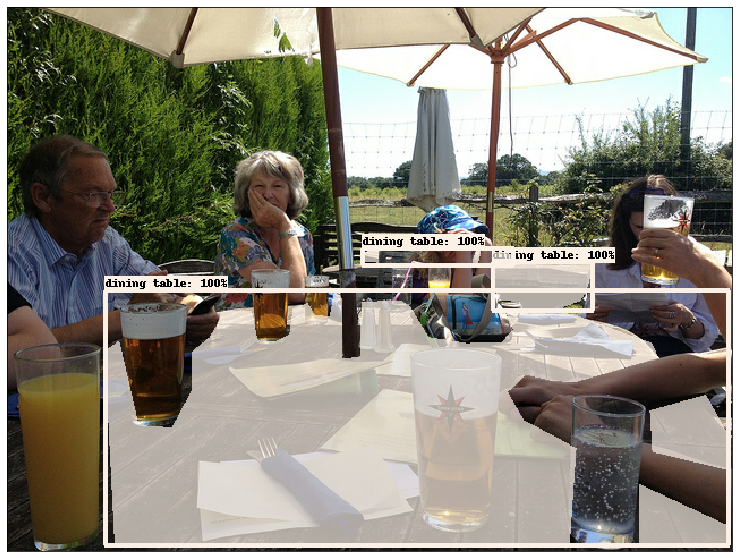}
\includegraphics[width=0.32\linewidth]{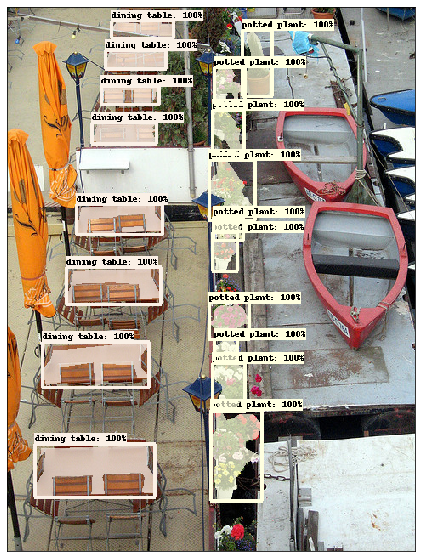}
\caption{Dining Table: Inconsistencies in annotated parts of dining tables.
Left: Plate and cup with carrots is excluded whereas plate with empty glass
is included in the mask. Center: Some glasses on the dining table are 
included as part of it whereas some classes aren't. Right: Chairs
are excluded from the dining table mask in the dining tables near the bottom,
whereas they are included in the dining table masks near the top.}
\end{subfigure}

\begin{subfigure}[t]{0.93\linewidth}
\includegraphics[width=0.32\linewidth]{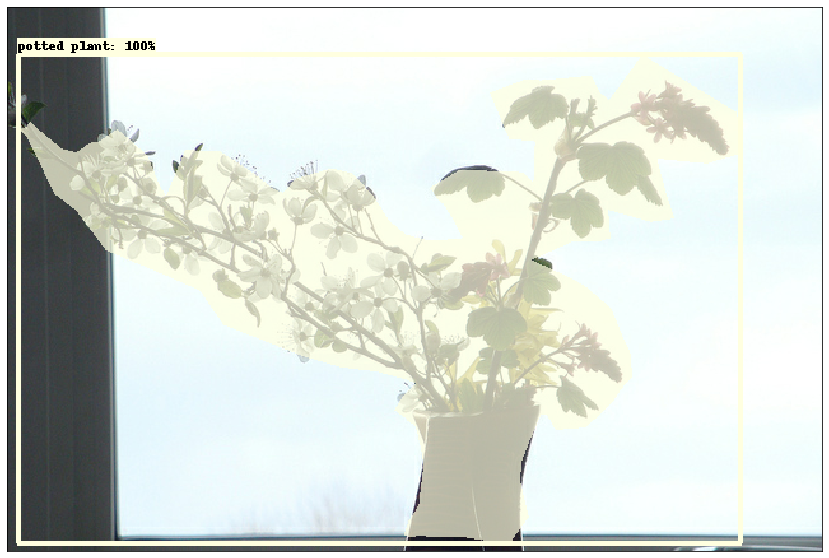}
\includegraphics[width=0.32\linewidth]{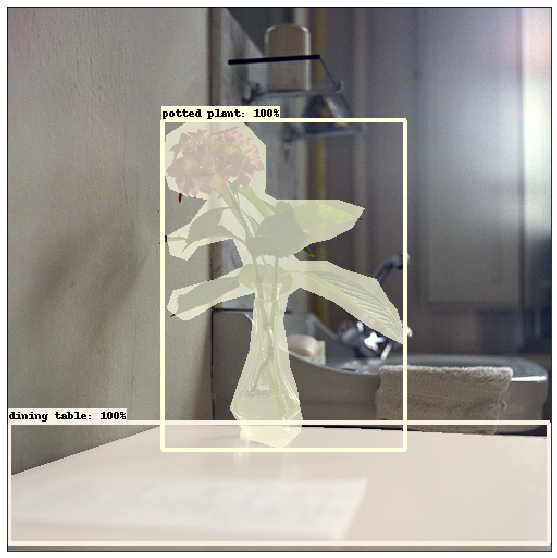}
\includegraphics[width=0.32\linewidth]{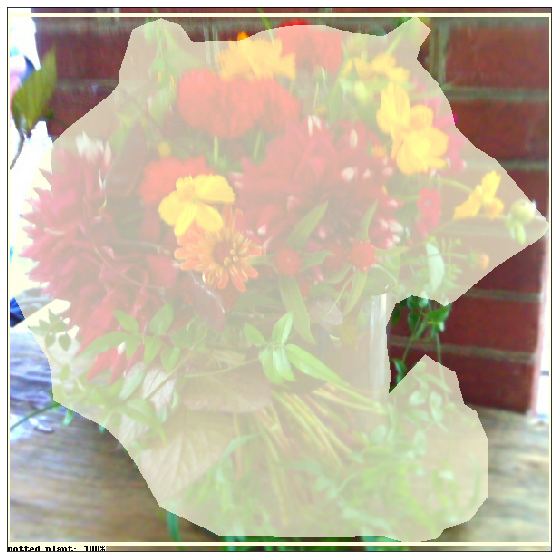}
\caption{Potted plant: Areas of background are included in the foreground
masks of potted plants, especially near the leaves.}
\end{subfigure}
\caption{Example annotations of the 3 worst source classes to train on.}
\label{fig:quality}
\end{figure*}

In Figure \ref{fig:quality} we show examples of COCO groundtruth annotations
from the dining table, bicycle and potted plant categories, the worst
three categories to use as source training categories.  The examples illustrate
the inconsistencies/inaccuracies in mask annotations for these categories --- for example,
annotators were inconsistent about including or excluding objects on the dining
tables.

\section{Mask head architecture details}
\label{sec:mask_head_details}
\begin{figure*}
    \centering
    \includegraphics[width=0.5\linewidth]{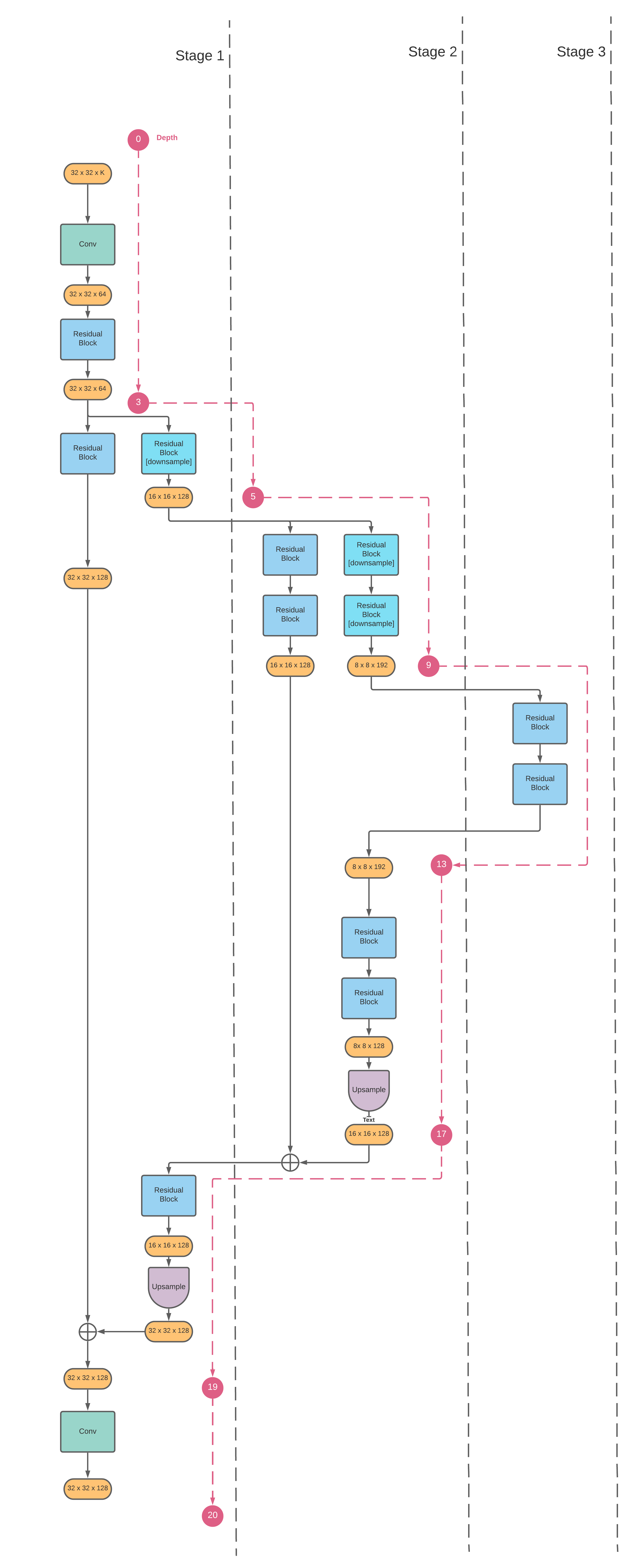}
    \caption{Illustration of the Hourglass 20 mask head computation graph.}
    \label{fig:hourglass_graph}
\end{figure*}

Details of mask head architectures can be found in Table \ref{tab:arch_details1}, \ref{tab:arch_details2}, \ref{tab:arch_details3} and \ref{tab:arch_details4}.
Figure \ref{fig:hourglass_graph} illustrates the computation graph of an hourglass mask head.
\begin{table*}[h]
\tablefontsize
\centering
\begin{tabular}{lrcrrr}
\toprule
{\bf Type} \hspace{30pt} & {\bf Depth}&\hspace{10pt} &{\bf \# of Blocks} & \multicolumn{2}{c}{\bf Conv Block}  \\
\cmidrule{5-6}
 & & & &{\bf Size} & {\bf Channels} \\
\midrule
ResNet & 4& & 1 & $32 \times 32$ & 64 \\
\cmidrule{4-6} 
&&&2 & $32 \times 32$ & 128 \\
&&& & $32 \times 32$ & 128 \\
\cmidrule{2-6}
 & 8& & 1 & $32 \times 32$ & 64 \\
\cmidrule{4-6} 
&&&4 & $32 \times 32$ & 128 \\
&&& & $32 \times 32$ & 128 \\
\cmidrule{2-6}
 & 12& & 1 & $32 \times 32$ & 64 \\
\cmidrule{4-6} 
&&& 6 & $32 \times 32$ & 128 \\
&&& & $32 \times 32$ & 128 \\
\cmidrule{2-6}
 & 16& & 1 & $32 \times 32$ & 64 \\
\cmidrule{4-6} 
&&& 8 & $32 \times 32$ & 128 \\
&&& & $32 \times 32$ & 128 \\
\cmidrule{2-6}
 & 20& & 1 & $32 \times 32$ & 64 \\
\cmidrule{4-6} 
&&& 8 & $32 \times 32$ & 128 \\
&&& & $32 \times 32$ & 128 \\
\cmidrule{4-6} 
&&& 2 & $32 \times 32$ & 128 \\
&&& & $32 \times 32$ & 128 \\
\cmidrule{1-6}
ResNet Bottleneck & 6& & 1 & $32 \times 32$ & 64 \\
\cmidrule{4-6} 
&&&2 & $32 \times 32$ & 128 \\
&&& & $32 \times 32$ & 512 \\
&&& & $32 \times 32$ & 128 \\
\cmidrule{2-6}
 & 9& & 1 & $32 \times 32$ & 64 \\
\cmidrule{4-6} 
&&&3 & $32 \times 32$ & 128 \\
&&& & $32 \times 32$ & 512 \\
&&& & $32 \times 32$ & 128 \\
\cmidrule{2-6}
 & 12& & 1 & $32 \times 32$ & 64 \\
\cmidrule{4-6} 
&&&4 & $32 \times 32$ & 128 \\
&&& & $32 \times 32$ & 512 \\
&&& & $32 \times 32$ & 128 \\
\cmidrule{2-6}
 & 15& & 1 & $32 \times 32$ & 64 \\
\cmidrule{4-6} 
&&&5 & $32 \times 32$ & 128 \\
&&& & $32 \times 32$ & 512 \\
&&& & $32 \times 32$ & 128 \\
\cmidrule{2-6}
 & 21& & 1 & $32 \times 32$ & 64 \\
\cmidrule{4-6} 
&&&6 & $32 \times 32$ & 128 \\
&&& & $32 \times 32$ & 512 \\
&&& & $32 \times 32$ & 128 \\
\cmidrule{4-6} 
&&&1 & $32 \times 32$ & 192 \\
&&& & $32 \times 32$ & 384 \\
&&& & $32 \times 32$ & 192 \\
\bottomrule
\end{tabular}
\caption{Architecture details of ResNet and ResNet bottleneck mask heads.}
\label{tab:arch_details1}
\end{table*}

\begin{table*}[h]
\tablefontsize
\centering
\begin{tabular}{lrcrrr}
\toprule
{\bf Type} \hspace{30pt} & {\bf Depth}&\hspace{10pt} &{\bf \# of Blocks} & \multicolumn{2}{c}{\bf Conv Block}  \\
\cmidrule{5-6}
 & & & &{\bf Size} & {\bf Channels} \\
\midrule
ResNet Bottleneck [1/4$^\mathrm{th}$] & 6& & 1 & $32 \times 32$ & 16 \\
\cmidrule{4-6} 
&&&2 & $32 \times 32$ & 32 \\
&&& & $32 \times 32$ & 128 \\
&&& & $32 \times 32$ & 32 \\
\cmidrule{2-6}
\cmidrule{2-6}
 & 12& & 1 & $32 \times 32$ & 16 \\
\cmidrule{4-6} 
&&&4 & $32 \times 32$ & 32 \\
&&& & $32 \times 32$ & 128 \\
&&& & $32 \times 32$ & 32 \\
\cmidrule{2-6}
 & 21& & 1 & $32 \times 32$ & 16 \\
\cmidrule{4-6} 
&&&6 & $32 \times 32$ & 32 \\
&&& & $32 \times 32$ & 128 \\
&&& & $32 \times 32$ & 32 \\
\cmidrule{4-6} 
&&&1 & $32 \times 32$ & 48 \\
&&& & $32 \times 32$ & 192 \\
&&& & $32 \times 32$ & 48 \\
\cmidrule{2-6} 
 & 30& & 1 & $32 \times 32$ & 16 \\
\cmidrule{4-6} 
&&&5 & $32 \times 32$ & 32 \\
&&& & $32 \times 32$ & 128 \\
&&& & $32 \times 32$ & 32 \\
\cmidrule{4-6} 
&&&5 & $32 \times 32$ & 48 \\
&&& & $32 \times 32$ & 192 \\
&&& & $32 \times 32$ & 48 \\
\cmidrule{2-6} 
\cmidrule{2-6} 
 & 51& & 1 & $32 \times 32$ & 16 \\
\cmidrule{4-6} 
&&&6 & $32 \times 32$ & 32 \\
&&& & $32 \times 32$ & 128 \\
&&& & $32 \times 32$ & 32 \\
\cmidrule{4-6} 
&&&8 & $32 \times 32$ & 48 \\
&&& & $32 \times 32$ & 192 \\
&&& & $32 \times 32$ & 48 \\
\cmidrule{4-6} 
&&&3 & $32 \times 32$ & 64 \\
&&& & $32 \times 32$ & 256 \\
&&& & $32 \times 32$ & 64 \\
\bottomrule
\end{tabular}
\caption{Architecture details of ResNet bottleneck [1/4$^\mathrm{th}$] mask head.}
\label{tab:arch_details2}
\end{table*}

\begin{table*}[h]
\tablefontsize
\centering
\begin{tabular}{lrcrrr}
\toprule
{\bf Type} \hspace{30pt} & {\bf Depth}&\hspace{10pt} &{\bf \# of Blocks} & \multicolumn{2}{c}{\bf Conv Block}  \\
\cmidrule{5-6}
 & & & &{\bf Size} & {\bf Channels} \\
\midrule
Hourglass & 10 && 1 & $32 \times 32$ &  64\\
\cmidrule{4-6}
&&& 3 & $ 32 \times 32$ & 128\\
&&& & $ 32 \times 32$ & 128\\
\cmidrule{4-6}
&&& 1 & $ 16 \times 16$ & 128\\
&&& & $ 16 \times 16$ & 128\\
\cmidrule{4-6}
& && 1 & $32 \times 32$ &  128\\
\cmidrule{2-6}
 & 20 && 1 & $32 \times 32$ &  64\\
\cmidrule{4-6}
&&& 3 & $ 32 \times 32$ & 128\\
&&& & $ 32 \times 32$ & 128\\
\cmidrule{4-6}
&&& 4 & $ 16 \times 16$ & 128\\
&&& & $ 16 \times 16$ & 128\\
\cmidrule{4-6}
&&& 2 & $ 8 \times 8$ & 192\\
&&& & $ 8 \times 8$ & 192\\
\cmidrule{4-6}
& && 1 & $32 \times 32$ &  128\\
\cmidrule{2-6}
 & 32 && 1 & $32 \times 32$ &  64\\
\cmidrule{4-6}
&&& 5 & $ 32 \times 32$ & 128\\
&&& & $ 32 \times 32$ & 128\\
\cmidrule{4-6}
&&& 4 & $ 16 \times 16$ & 128\\
&&& & $ 16 \times 16$ & 128\\
\cmidrule{4-6}
&&& 4 & $ 8 \times 8$ & 192\\
&&& & $ 8 \times 8$ & 192\\
\cmidrule{4-6}
&&& 2 & $ 4 \times 4$ & 192\\
&&& & $ 4 \times 4$ & 192\\
\cmidrule{4-6}
& && 1 & $32 \times 32$ &  128\\
\cmidrule{2-6}
 & 52 && 1 & $32 \times 32$ &  64\\
\cmidrule{4-6}
&&& 5 & $ 32 \times 32$ & 128\\
&&& & $ 32 \times 32$ & 128\\
\cmidrule{4-6}
&&& 4 & $ 16 \times 16$ & 128\\
&&& & $ 16 \times 16$ & 128\\
\cmidrule{4-6}
&&& 4 & $ 8 \times 8$ & 192\\
&&& & $ 8 \times 8$ & 192\\
\cmidrule{4-6}
&&& 4 & $ 4 \times 4$ & 192\\
&&& & $ 4 \times 4$ & 192\\
\cmidrule{4-6}
&&& 4 & $ 2 \times 2$ & 192\\
&&& & $ 2 \times 2$ & 192\\
\cmidrule{4-6}
&&& 4 & $ 1 \times 1$ & 256\\
&&& & $ 1 \times 1$ & 256\\
\cmidrule{4-6}
& && 1 & $32 \times 32$ &  128\\
\bottomrule
\end{tabular}
\caption{Architecture details of Hourglass mask head (Part 1 of 2).}
\label{tab:arch_details3}
\end{table*}

\begin{table*}[h]
\tablefontsize
\centering
\begin{tabular}{lrcrrr}
\toprule
{\bf Type} \hspace{30pt} & {\bf Depth}&\hspace{10pt} &{\bf \# of Blocks} & \multicolumn{2}{c}{\bf Conv Block}  \\
\cmidrule{5-6}
 & & & &{\bf Size} & {\bf Channels} \\
\midrule
Hourglass & 100 && 1 & $32 \times 32$ &  64\\
\cmidrule{4-6}
&&& 9 & $ 32 \times 32$ & 128\\
&&& & $ 32 \times 32$ & 128\\
\cmidrule{4-6}
&&& 8 & $ 16 \times 16$ & 128\\
&&& & $ 16 \times 16$ & 128\\
\cmidrule{4-6}
&&& 8 & $ 8 \times 8$ & 192\\
&&& & $ 8 \times 8$ & 192\\
\cmidrule{4-6}
&&& 8 & $ 4 \times 4$ & 192\\
&&& & $ 4 \times 4$ & 192\\
\cmidrule{4-6}
&&& 8 & $ 2 \times 2$ & 192\\
&&& & $ 2 \times 2$ & 192\\
\cmidrule{4-6}
&&& 8 & $ 1 \times 1$ & 256\\
&&& & $ 1 \times 1$ & 256\\
\cmidrule{4-6}
& && 1 & $32 \times 32$ &  128\\
\bottomrule
\end{tabular}
\caption{Architecture details of Hourglass mask head (Part 2 of 2).}
\label{tab:arch_details4}
\end{table*}

\end{appendices}
\end{document}